\documentclass{article}

    \PassOptionsToPackage{numbers, compress}{natbib}

    \usepackage[preprint]{neurips_2024}

\usepackage[utf8]{inputenc} 
\usepackage[T1]{fontenc}    
\usepackage{hyperref}       
\usepackage{url}            
\usepackage{booktabs}       
\usepackage{amsfonts}       
\usepackage{nicefrac}       
\usepackage{microtype}      
\usepackage{xcolor}         
\usepackage{lipsum} 
\usepackage{hhline}
\usepackage{boldline}
\usepackage{multirow}
\usepackage{amsmath}
\usepackage{amssymb}
\usepackage{pifont}
\usepackage{enumitem}
\usepackage{caption}
\usepackage{graphicx} 
\usepackage{wrapfig}
\usepackage{algorithm,algorithmic}
\usepackage{colortbl}
\usepackage{CJKutf8} 
\usepackage{bm} 
\usepackage{bbm} 
\usepackage{footmisc} 
\usepackage{hyperref}
\usepackage{xcolor}
\usepackage{fontawesome}
\hypersetup{
colorlinks=true,
linkcolor=black
}
\definecolor{uclablue}{RGB}{50,132,191}

\newcommand{\Rmnum}[1]{\expandafter\@slowromancap\romannumeral #1@}

\newcommand{\RomanNumeralCaps}[1]{\MakeUppercase{\romannumeral #1}}

\newcommand*\diff{\mathop{}\!\mathrm{d}}

\newcommand{\qcr}[1]{{\fontfamily{cmtt}\selectfont #1}}
\newcommand{\nig}{\text{\qcr{NIG}}} 
\newcommand{\unit}{clip}
\newcommand\E{\mathbb{E}}
\newcommand{\Var}{\mathrm{Var}}

\newcommand{\Loss}{\mathcal{L}}
\newcommand{\Lnll}{\Loss^{\scalebox{.7}{{NLL}}}}

\title{Beyond Uncertainty: Evidential Deep Learning for Robust Video Temporal Grounding}
\author{%
  Kaijing Ma{$^{\spadesuit \heartsuit}$}\thanks{These authors contributed equally.} \ , \ Haojian Huang{$^{\heartsuit \clubsuit}$}\footnotemark[1], \ Jin Chen {$^{\spadesuit \heartsuit}$}\footnotemark[1], \ Haodong Chen{$^{\diamondsuit \heartsuit}$}, \\
  \textbf{Pengliang Ji$^{\star}$, \ Xianghao Zang{$^{\heartsuit}$}, \ Han Fang$^{\heartsuit}$,
  \ Chao Ban$^{\heartsuit}$}, \\
  \textbf{Hao Sun$^{\heartsuit}$\footnotemark[2], \ Mulin Chen{$^{\diamondsuit \heartsuit}$}\footnotemark[2], \ Xuelong Li$^{\heartsuit}$}\thanks{Correspondence to: \ \texttt{ 
    \{xjtumakaijing, haojianhuang927, chenmulin001\}@gmail.com, \\
    ~sun.010@163.com, ~li@nwpu.edu.cn
  }} \\
{$^\heartsuit$TeleAI}  \quad
{$^\spadesuit$Xi'an Jiaotong University}  \quad 
{$^\clubsuit$The University of Hong Kong} \\
{$^\diamondsuit$Northwestern Polytechnical University}  \quad
{$^\star$ Carnegie Mellon University} \\
\href{https://kaijing.space/SRAM/}{
  \faGithub\ \textcolor{uclablue}{\textbf{\ttfamily \small TeleAI SRAM}}
}
\vspace{-15pt}
}

\begin{document}
\renewcommand{\bibfont}{\small}

\maketitle

\begin{abstract}
Existing Video Temporal Grounding (VTG) models excel in accuracy but often overlook open-world challenges posed by open-vocabulary queries and untrimmed videos. This leads to unreliable predictions for noisy, corrupted, and out-of-distribution data. Adapting VTG models to dynamically estimate uncertainties based on user input can address this issue.
To this end, we introduce \textbf{SRAM}, a robust network module that benefits from a two-stage cross-modal alignment task. More importantly, it integrates Deep Evidential Regression (DER) to explicitly and thoroughly quantify uncertainty during training, thus allowing the model to say ``I do not know'' in scenarios beyond its handling capacity. However, the direct application of traditional DER theory and its regularizer reveals structural flaws, leading to unintended constraints in VTG tasks. In response, we develop a simple yet effective Geom-regularizer that enhances the uncertainty learning framework from the ground up. To the best of our knowledge, this marks the first successful attempt of DER in VTG. Our extensive quantitative and qualitative results affirm the effectiveness, robustness, and interpretability of our modules and the uncertainty learning paradigm in VTG tasks. The code will be made available.

\end{abstract}

\section{Introduction}
Video is emerging as the primary information carrier in the era of streaming media. With the influx of video data, the need for efficiently and precisely extracting desired content from videos according to users' queries is becoming increasingly essential.
In response to these demands, Video Temporal Grounding (VTG) emerges as a core research area in the field of computer vision~\cite{dosovitskiy2020image, li2023mvbench, chen2024gaussianvton, chen2024finecliper, huang2024crest}. Given an untrimmed video with a textual query, VTG can be categorized into three main types: 1) accurately identifying specific segments, referred to as \textit{Moment Retrieval}~\cite{gao2017tall, grauman2022ego4d, lei2021detecting, liu2022umt}; 2) comprehending the meaningful gist of a video, known as \textit{Highlight Detection}~\cite{lei2021detecting, sun2014ranking, xiong2019less, hong2020mini}; and 3) grasping the overall content of the entire video or specific sections, termed \textit{Video Summarization}~\cite{sharghi2017query, apostolidis2021video, gygli2014creating, song2015tvsum}. 

\begin{figure*}
\vspace{-0.3cm}
 \centering
\includegraphics[width=0.95\linewidth]{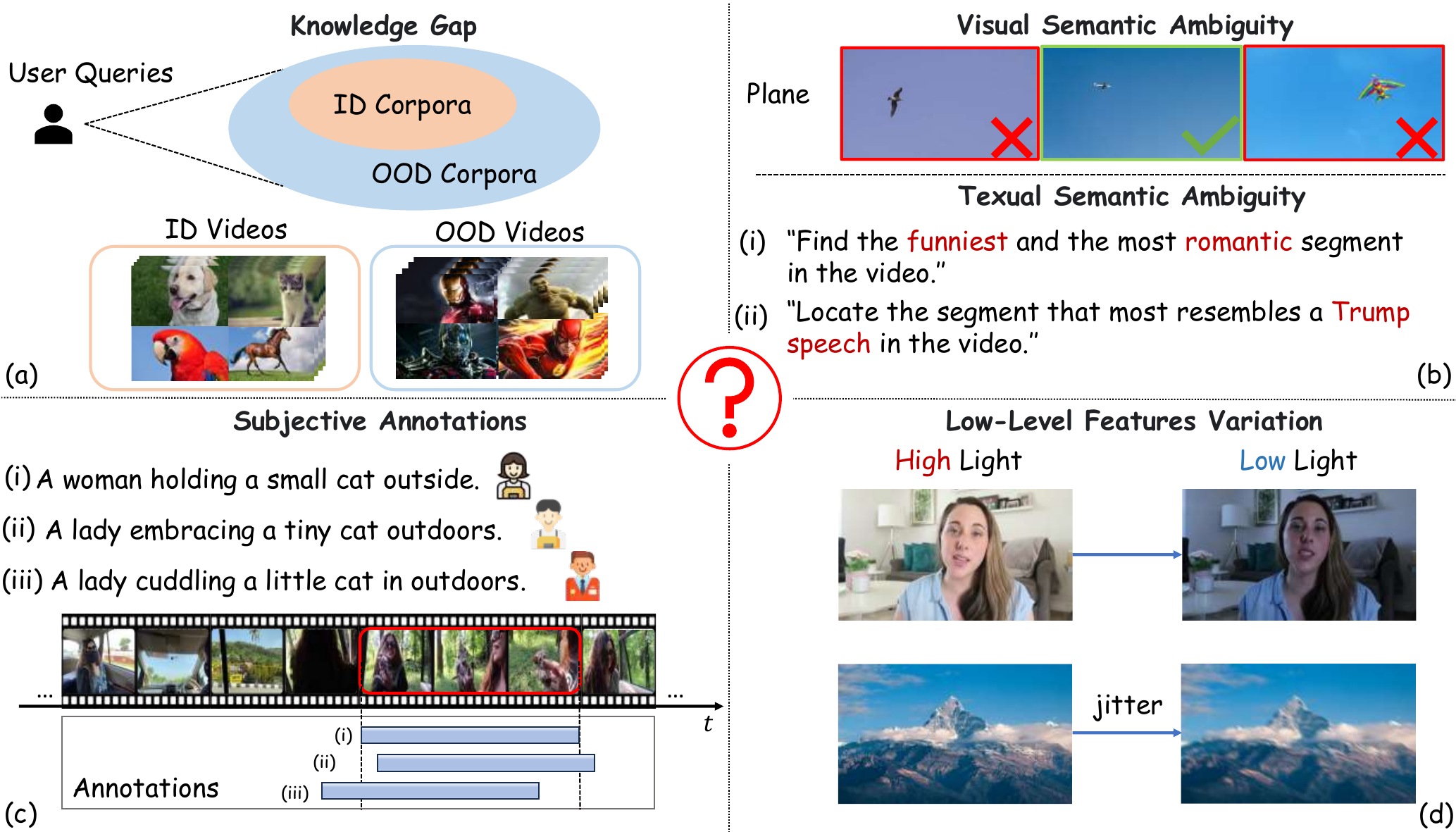}
\vspace{-0.1cm}
\caption{\textbf{Motivation illustration.} Epistemic uncertainty arises mainly from Knowledge Gaps and Semantic Ambiguity in (a) and (b), while aleatoric uncertainty is primarily due to Subjective Annotations and Low-Level Feature Uncertainty in (c) and (d). Specifically, (a) illustrates a critical knowledge gap where the model’s training does not sufficiently cover the possible real-world scenarios, leading to potential failures in understanding and responding to user inputs, while (b) highlights the challenges of semantic ambiguity in both visual and textual contexts. (c) indicates the subjective nature of annotations in datasets, while (d) unfolds the potential challenge led by variations in scene lighting, resolution, jittering, blurring, and transitions \emph{etc}. Exemplification can be found in Appendix \ref{ad_noise} and \ref{cases_study}.}
\label{motivation_fig}  
\vspace{-6pt}
\end{figure*}

Extensive research aimed at enhancing cross-modal reasoning to facilitate fine-grained and precise multi-modal alignment has led to significant advances in the field of VTG~\cite{lin2023univtg}. Nevertheless, few studies have focused on the epistemic and aleatoric uncertainties present in open-world human-computer interaction (HCI)~\cite{amini2020deep}. To begin with, epistemic uncertainty can be attributed to Knowledge Gap and Semantic Ambiguity as shown in Figure~\ref{motivation_fig} (a) and (b) respectively. Specifically, user queries and video inputs often come from out-of-distribution (OOD) sources, diverging significantly from the in-distribution (ID) data used in training. This natural discrepancy creates a Knowledge Gap, making it challenging for models to accurately understand and respond to user needs. Moreover, semantic ambiguities hinder accurate contextual understanding. Visual ambiguities often occur in images with extensive uniform backgrounds, such as planes in the sky, where the distinction between foreground and background blurs. On the textual side, ambiguities stem from subjective elements like emotions or speaking styles, making the intended meaning open to various interpretations.
While epistemic uncertainty measures the uncertainty in model inference due to incomplete knowledge, aleatoric uncertainty arises from inherent variability in the dataset. 
\begin{wrapfigure}{r}{0.5\textwidth}
\vspace{-0.3cm}
    \centering
    \includegraphics[width=0.5\textwidth]{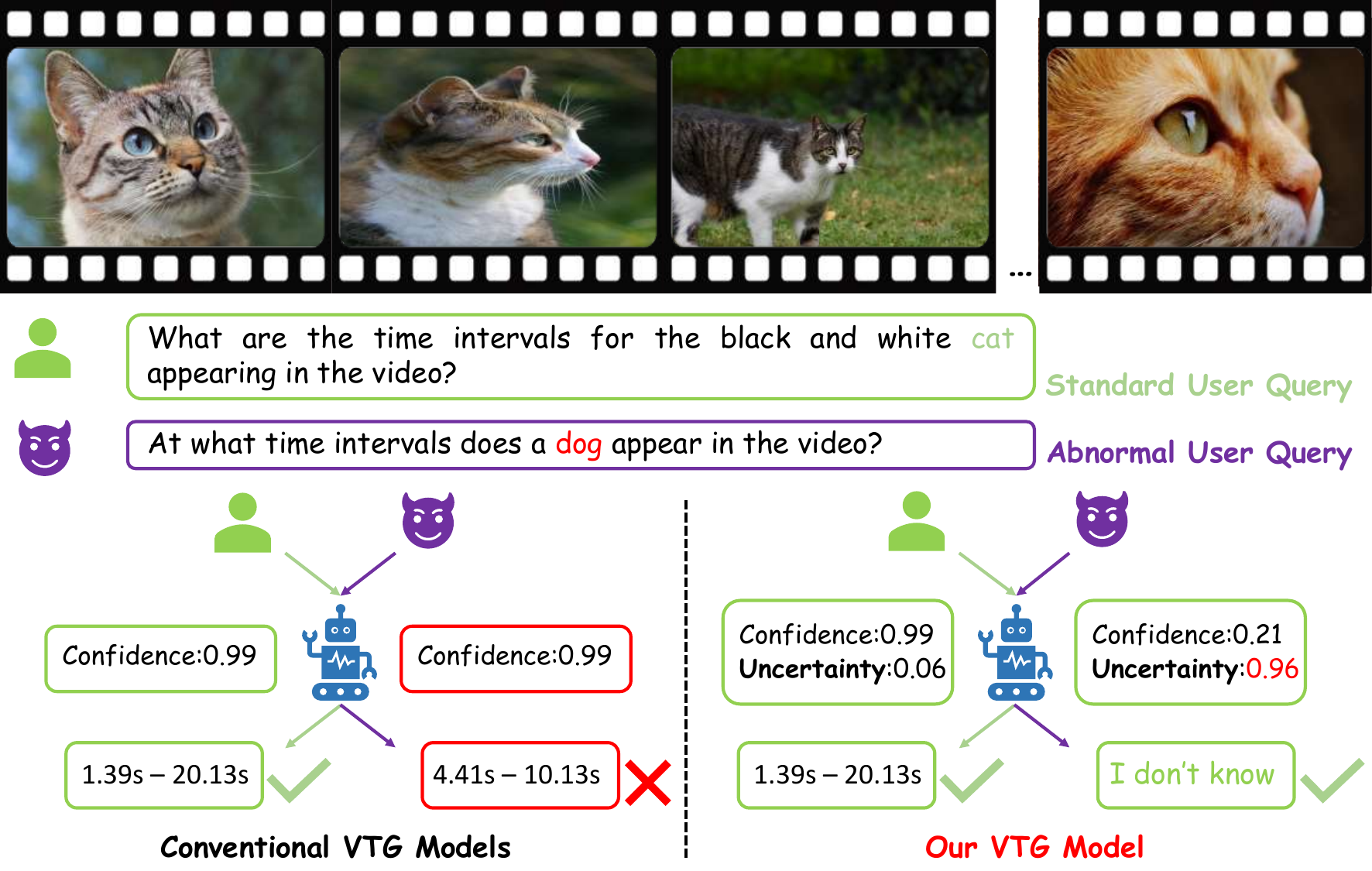}
    \vspace{-0.1cm}
    \caption{Conceptual illustration: Conventional models give random responses to OOD queries, unfit for critical decisions. In contrast, our model reliably delivers sensible, informed answers.}
    \label{comp}
\end{wrapfigure}
In VTG, aleatoric uncertainty typically arises from subjective annotations and variations in low-level features. Subjective annotations occur when different annotators provide varying queries and labels for the same sample, influenced by their personal views and habits. Additionally, randomness stems from differences in video features such as texture, edges, resolution, lighting, camera jittering, and scene transitions, all of which disrupt the consistency of video analysis. Unfortunately, when confronted with anomalies, such as the abnormal queries shown in Figure~\ref{comp}, conventional VTG models often respond with nearly random and indiscreet answers. These models fail to handle potential uncertainties in an open world appropriately. This inadequacy is detrimental in scenarios that require cautious decision-making, such as security and confidential environments.

To this end, we attempt to integrate deep evidential regression (DER)~\cite{amini2020deep} in VTG for the first time, leading to a novel network module, named \underline{S}crupulous \underline{R}efinement with uncertainty-\underline{A}wareness \underline{M}odule, termed as SRAM. Specifically, SRAM incorporates reflective flipped fusion (RFF) blocks and achieves fine-grained cross-modal alignment through a two-stage cross-modal alignment task. Additionally, by introducing an evidential head, it comprehensively and explicitly measures aleatoric and epistemic uncertainties in VTG. This allows the model to progressively optimize its understanding while clearly grasping its capability boundaries, yielding trustworthy performance. Unfortunately, the vanilla DER uses the product of error and evidence as a regularizer. With prolonged training, the model tends to overly suppress evidence. This leads to significant bias in the model's uncertainty reasoning. We argue that this stems from the structure flaws in the regularizer. To address this, we propose a simple yet effective Geom-regularizer, which ensures the model yields sensible uncertainty and robustly adapts to open-world HCI.

To summarize, our main contributions are three-fold: \textbf{1)} We address the challenge of uncertainties in VTG by employing DER to explicitly and comprehensively measure potential uncertainties for the first time. Additionally, we design a simple yet effective Geom-regularizer to correct structural flaws in the vanilla DER regularizer, enhancing the model's robustness and trustworthiness in open world. \textbf{2)} We propose a two-stage cross-modal alignment task and employ the RFF blocks for progressive fine-grained alignment of text and video. \textbf{3)} Extensive experiments demonstrate that our proposed SRAM exhibit effectiveness, robustness, and interpretability across multiple benchmarks.

\section{Related work}
\subsection{Video temporal grounding}
Video Temporal Grounding (VTG) identifies correlated shots in videos based on natural language queries, which broadly supports various downstream video comprehension tasks, such as video moment retrieval~\cite{anne2017localizing,chen2018temporally,zhang2020learning,moon2023query,li2024momentdiff}, highlight detection~\cite{rui2000automatically,sun2014ranking,moon2023query}, and video summarization~\cite{gygli2014creating,jiang2022joint,mahasseni2017unsupervised,nalla2020watch,sharghi2017query,wu2022intentvizor}. These tasks generally involve formulating the boundaries of significant semantic segments~\cite{jiang2022joint,moon2023query,lin2023univtg}. Numerous innovative and effective methods have been developed to address the challenges in VTG. For instance, CTRL~\cite{anne2017localizing} and MCN~\cite{gao2017tall} initially generate proposals using sliding windows, which rank in terms of a cross-modal matching score. MomentDETR~\cite{lei2021detecting} applies a transformer to predict potential moments through learnable queries. Furthermore, QD-DETR~\cite{moon2023query} employs a cross-attention module and a negative pair training scheme to enhance multi-modal alignment. MomentDiff~\cite{li2024momentdiff} initially sets random boundaries for temporal segments and iteratively refines them to better match the intended semantics. However, existing approaches typically yield deterministic predictions, operating under the assumption that semantic segments are demarcated by clear and precise boundaries. This presumption neglects the inherent ambiguity and uncertainty associated with determining the true extent of these segments. To address this gap, we explicitly model and quantify the semantic uncertainty of video segment boundaries.

\subsection{Uncertainty learning}
Recent studies have highlighted inherent ambiguities and biases in VTG datasets, which significantly impact the integrity and performance of models~\cite{zhou2021embracing, zhang2023temporal}. These uncertainties are categorized into annotation and query uncertainties. Annotation uncertainty stems from varying temporal boundaries assigned by different annotators to the same query, while query uncertainty arises from the use of differing descriptions for the same video moment, underscoring the subjective nature of video interpretation~\cite{zhou2021embracing, zhang2023temporal}. Furthermore, these datasets exhibit pronounced biases, with common events being overly represented and a small subset of queries accounting for most actions, leading to a skewed distribution that creates a long tail in ground-truth timestamps~\cite{zhang2023temporal, otani2020uncovering}. These findings underscore the need for meticulous curation of datasets and the adoption of uncertainty-aware modeling approaches~\cite{arnab2020uncertainty, malinin2018predictive, zhou2021amortized, gawlikowski2023survey}. Among the techniques for modeling uncertainty, Evidential Deep Learning (EDL) has shown promise. Originating from the principles of Dempster-Shafer Theory~\cite{shafer1992dempster} and Subjective Logic~\cite{sensoy2018evidential, josang2016subjective}, EDL models uncertainty explicitly through the distribution of "second-order probabilities" over network outputs, finding applications across various classification tasks including action recognition~\cite{bao2021evidential}, multi-view clustering~\cite{han2020trusted,han2022trusted}, zero-shot learning~\cite{huang2024crest}, and semantic segmentation~\cite{holmquist2023evidential} \emph{etc}. Leveraging DER~\cite{amini2020deep} for its extension, EDL has effectively been applied to regression tasks such as stereo matching~\cite{wang2022uncertainty} and emotion attributes estimation~\cite{wu2023estimating}.
Nevertheless, DER faces challenges like evidence contraction due to the non-negativity of prior parameters in the Normal Inverse-Gamma (\nig) distribution. New regularizers have been developed to address these issues, enhancing reliability and performance~\cite{wu2024evidence}. However, DER often encounters gradient disappearance in high uncertainty areas, necessitating ongoing refinement of its regularization methods~\cite{ye2024uncertainty, meinert2023unreasonable}. In this study, we apply DER within the VTG framework to manage uncertainties in open-world user inputs. We also identify and improve the structural flaws of the vanilla DER regularizer. To the best of our knowledge, this marks the first successful attempt to integrate DER in an uncertainty-aware VTG framework.

\begin{figure*}
\vspace{-0.3cm}
 \centering
\includegraphics[width=\linewidth]{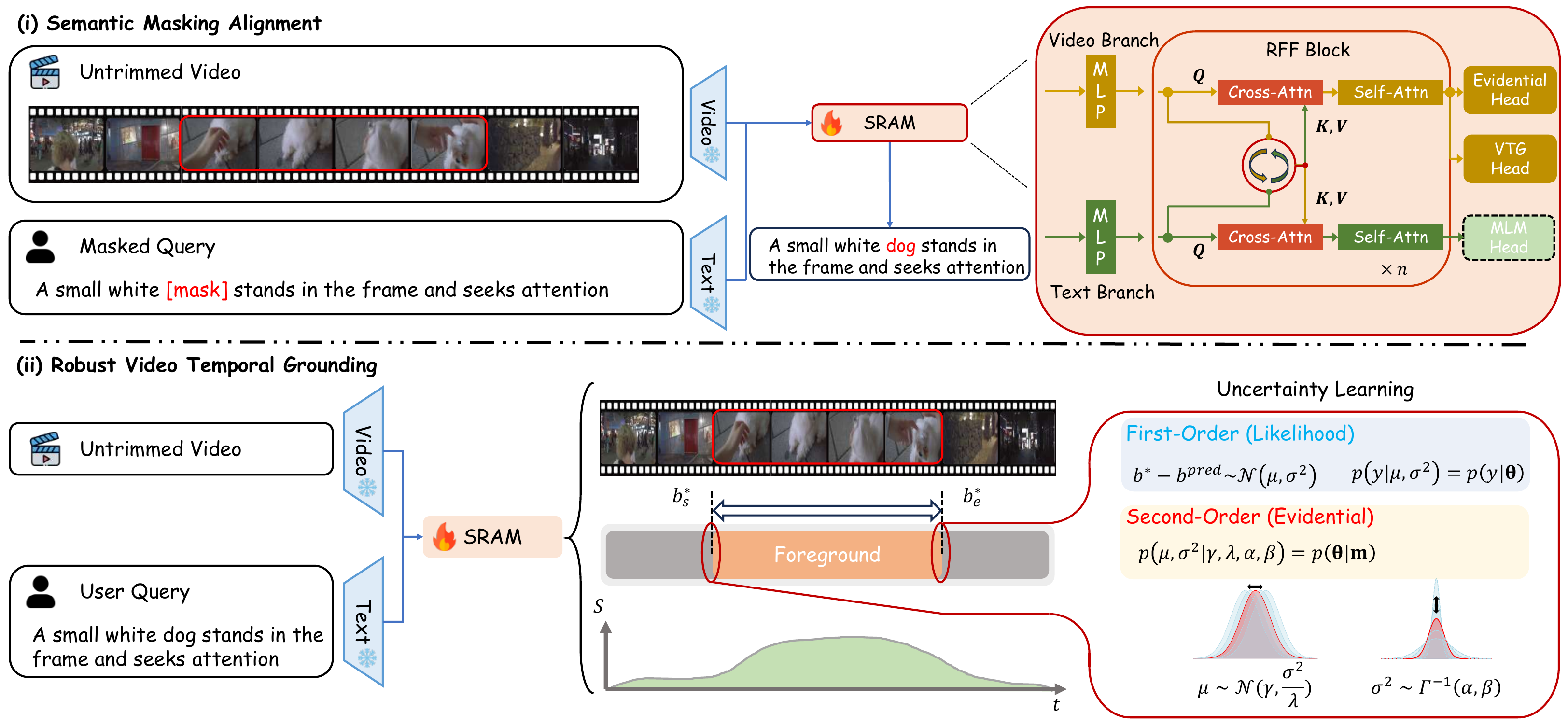}
\vspace{-0.3cm}
\caption{Overall architecture of the proposed two-stage cross-modal alignment using SRAM. Firstly, an untrimmed video and masked query are encoded with a frozen encoder, then SRAM reconstructs the masked query tokens. In the second stage, SRAM performs temporal grounding on the video using the complete user's query. SRAM includes RFF blocks, an evidential head, a VTG head, and a Masked Language Model (MLM) head. The MLM head enclosed by the dashed box is trained only during the first stage. These components are discussed in the following sections.}

\label{overview}  
\vspace{-5pt}
\end{figure*}
\section{Methodology}
\subsection{Problem definition}
\label{problem}
Given a video \( V = \{v_i\}_{i=1}^{L_v} \) and a query \( Q = \{q_i\}_{i=1}^{L_q} \), each represented as vectors in \( \mathbb{R}^{D} \) where \( L_v \) and \( L_q \) denote the counts of clips and tokens respectively, the task of VTG is to assign each clip a label \( (m_i, s_i, f_i) \) signifying its relevance to \( Q \).
\begin{itemize}
    \item \textbf{Moment Retrieval}: Identify clips in \( V \) corresponding to moments in \( Q \), with $m_i = [m^s_i, m^e_i]$ marking the time span of the nearest relevant moment.
    \item \textbf{Highlight Detection}: Determine each clip's relevance to \( Q \) using a saliency score \( s_i \), which reflects semantic alignment, scaled between [0, 1].
    \item \textbf{Video Summarization}: Select clips for a concise video summary, where \( f_i \) indicates whether a clip is included, taking values in \{0, 1\}.
\end{itemize}

\vspace{-0.5cm}
\subsection{Overview}
Figure~\ref{overview} illustrates our method's workflow, starting with encoding an untrimmed video and masked query using a frozen encoder. Our process involves two stages. Firstly, SRAM reconstructs masked query tokens with RFF blocks for cross-modal alignment. Next, it performs VTG based on the query. The evidential head assesses aleatoric and epistemic uncertainties, while the VTG and MLM heads manage corresponding task stages. Details on these components are in the subsequent sections.
\vspace{-0.3cm}
\subsection{Scrupulous reflection with uncertainty awareness module} 
\textbf{Reflective flipped fusion block.}
The Reflective Flipped Fusion (RFF) block processes inputs from the video and text branches, alternating the roles of video and text as queries and keys/values using shared parameters. Through the cross-attention module, initial features $V^{(1)}$ and $Q^{(1)}$ of video and text branches are respectively updated, reflecting each other's information. Following cross-attention, each branch refines its features through self-attention to enhance internal feature representation. The outputs of self-attention serve as inputs to the next iteration of the RFF block, progressively enhancing modal alignment. This process is applied sequentially from block 1 to block $n$ in terms of Eq.~\ref{RFF}. After completing the $n$-th layer, the refined video and query features are output. The specific workflow process is detailed in Appendix~\ref{sec:workflow}.

\begin{equation}
\label{RFF}
V^{(i+1)} = SA_v^{(i)}(CA_{q \to v}^{(i)}), \quad Q^{(i+1)} = SA_q^{(i)}(CA_{v \to q}^{(i)}), \quad i = 1, 2, \dots, n-1
\end{equation}

\textbf{Evidential head}.
Temporal continuity in videos often causes adjacent frames to share similar semantics, complicating precise boundary delineation and introducing subjective biases in annotations. To mitigate this, we model semantic boundary moments using Gaussian distributions. Specifically, the start and end moments of a video-query pair \( (V, Q) \) are each governed by distinct Gaussian distributions. Observations of the same type (either all starts or all ends) are assumed to be \emph{i.i.d.}. Without loss of generality, we formulate as follows:
\begin{equation}
 \bm b \sim\mathcal{N}(\mu,\sigma^2),
\end{equation}
where $\bm b \in \mathbb{R}^{1 \times \mathcal{H}}$ represents the start or end of moments observed $\mathcal{H}$ times. The corresponding expectation $\mu$ and variance $\sigma^2$ of the Gaussian distribution subject to \nig~prior:
\begin{align}
p(\mu,\sigma^2\mid\underbrace{\gamma,\upsilon,\alpha,\beta}_{\boldsymbol{\varphi }})
&=
\mathcal{N}(\mu|\gamma,\sigma^2 \upsilon^{-1})
\Gamma^{-1}(\sigma^2|\alpha,\beta) \\
&=\frac{\beta^\alpha\sqrt{\upsilon}}{\Gamma(\alpha)\sqrt{2\pi\sigma^2}}\left(\frac{1}{\sigma^2}\right)^{\alpha+1}
\exp\left\{-\frac{2\beta+\upsilon(\gamma-\mu)^2}{2\sigma^2}\right\},
\end{align}
where $\boldsymbol{\varphi}=(\gamma, \upsilon, \alpha, \beta)$ are the prior \nig~distribution parameters derived from the video content and user queries, serve as conditionals for the Gaussian estimates of \( b_i \), with $\gamma \in \mathbb{R}, \upsilon > 0, \alpha > 1, \beta > 0$. The gamma function is denoted by $\Gamma(\cdot)$. We use a linear evidential predictor to estimate $\boldsymbol{\varphi }$, training it to maximize the likelihood. The maximum likelihood estimation for $b_i$ is given by:
\begin{equation}
p(b_i \mid \boldsymbol{\varphi }) = \int_{\sigma^2=0}^\infty \int_{\mu=-\infty}^\infty p(b_i \mid \mu, \sigma^2) p(\mu, \sigma^2 \mid \boldsymbol{\varphi }) d\mu d\sigma^2 = \mathrm{St}(b_i; \gamma, \frac{\beta(1 + \upsilon)}{\upsilon \alpha}, 2\alpha).
\end{equation}
Since the likelihood function has a form of Student-t distribution ($\mathrm{St}$), we minimize the negative logarithmic likelihood (NLL) as follows. Detailed formulation can be found in Appendix~\ref{sec:sup-nig-derivations} and~\ref{sec:sup-loss-derivations}.
\begin{equation}
\label{NLL_Loss}
\mathcal{L}^{\mathrm{NLL}}_{i}=
-\log p(b_i|\boldsymbol{\varphi })=
-\log\left(\mathrm{St}\left(b_i;\gamma,\frac{\beta(1+\upsilon)}{\upsilon\alpha},2\alpha\right)\right).
\end{equation}
Using the \nig~distribution, prediction, aleatoric, and epistemic uncertainties are calculated as follows:
\begin{equation}
\underbrace{\mathbb{E}[\mu] = \gamma}_{\text{prediction}},\quad\underbrace{\mathbb{E}[\sigma^2]=\frac{\beta}{\alpha-1}}_{\text{aleatoric}}, \quad\underbrace{\mathrm{Var}[\mu]=\frac{\beta}{\upsilon(\alpha-1)}}_{\text{epistemic}}.
\end{equation}
For guiding evidential preditor to evaluate more reasonable and informative uncertainty, we attempt to design Geom-regularization to assist training. We introduce it in detail in Section~\ref{geom}.

\textbf{Video temporal grounding head}.
The VTG head features three distinct modules for tasks outlined in section~\ref{problem}. For Video Summarization, the output from the frozen video encoder undergoes three 1x3 Conv layers, each with a ReLU activation. The Moment Retrieval head is similar but outputs two channels for offsets. Highlight Detection uses attentive pooling to form a sentence representation from query tokens, then computes the saliency score between video tokens and query as their cosine similarity. Details for each module and corresponding loss are available in the Appendix~\ref{VTG}.

\textbf{Semantic masking alignment}.
To ensure robust cross-modal alignment capabilities, during the initial phase of alignment, entities within the query are masked at a specified ratio. This approach compels the model to leverage contextual information available from the corresponding video and the remaining unmasked tokens in the query. Through the MLM head, the model infers and reconstructs the masked tokens. The loss function associated with this process, aimed at optimizing the model's cross-modal inference capabilities, is outlined below:
\begin{equation}
\mathcal{L}_{mlm} = \mathbb{E} \left[ -\sum_{i=1}^{l} \log P(w_i \mid U, V) \right],
\end{equation}
where \( l \) represents the number of masked tokens, \( w_i \) the \( i \)-th masked token, \( U \) the unmasked tokens providing linguistic context, and \( V \) the video features that enhance cross-modal contextual understanding for accurate token prediction. After the warm-up in the first phase, in the next phase, the MLM head is frozen and \(\mathcal{L}_{\text{mlm}}\) is not computed.

\subsection{Geom-regularization}
\label{geom}
Models optimized only on observed samples with the NLL loss (\emph{i.e.} Eq.~\ref{NLL_Loss}) tend to overfit and exhibit overconfidence. To counter this, DER introduced a regularizer for the $i$-th prediction as follows:
\begin{equation}
\label{vallina_reg}
\mathcal{L}^\mathrm{R}_i(\boldsymbol{\vartheta})=\Delta\cdot\Phi,
\end{equation}
where $\Delta = |b_i-\gamma|$ represents the error, $\Phi = 2\upsilon+\alpha$ denotes the evidence, and $\boldsymbol{\vartheta}$ are the model parameters, with $b_i$ as the ground truth. This heuristic regularization aims to mitigate overconfidence by suppressing evidence, particularly for samples with high error. However, excessive suppression can lead to underconfidence due to non-adaptive suppression and sample imbalance. To be clear, we first consider the minus gradient of $\mathcal{L}^\mathrm{R}_i$ with respect to $\Phi$ as follows:
\begin{equation}
-\nabla_{\Phi}\mathcal{L}^\mathrm{R}_i = - \Delta,
\end{equation}
\textbf{Non-adaptive suppression.}
We find that the gradient is solely related to error and unrelated to evidence, which means the model cannot determine when evidence has been sufficiently suppressed, as shown in Appendix~\ref{grad}. This can lead to excessively harsh penalties on evidence.

\textbf{Penalties bias.}
As the model converges, the dominance of low error samples with small gradients skews the batch's average gradient. Consequently, this leads to over-suppression of their evidence, while high error samples see their evidence neglected or adversely adjusted, as shown in Appendix~\ref{sec:evo}.

To overcome these limitations, we introduce Geom-regularization, inspired by~\cite{amini2020deep}, promoting the principle that "\textbf{\textit{accurate predictions should have high evidence, while inaccurate ones should have low evidence}}". This approach provides more rational constraints rather than merely suppressing evidence. Initially, we normalize $\Delta$ to $\overline{\Delta}$ and $\Phi$ to $\overline{\Phi}$ (\emph{i.e.} Appendix~\ref{norm}), which ensures that the model assigns $\overline{\Phi}=1$ to samples with $\overline{\Delta}=0$, and $\overline{\Phi}=0$ to samples with $\overline{\Delta}=1$.
We then ensure that the points $(\overline{\Delta},\overline{\Phi})$ closely follow the line $\overline{\Phi}+\overline{\Delta} = 1$ using a Type \RomanNumeralCaps{1}-line regularizer as below:
\begin{equation}
\label{loss:type1}
\mathcal{L}^\text{\RomanNumeralCaps{1}-L}_i(\boldsymbol{w})=\|\overline{\Phi}+\overline{\Delta}-1\|^2_2,
\end{equation}
we can follow the analysis for $\mathcal{L}^\mathrm{R}_i$. The minus gradient of $\mathcal{L}^\text{\RomanNumeralCaps{1}-L}_i$ with respect to $\overline{\Phi}$ as below:
\begin{equation}
-\nabla_{\overline{\Phi}}\mathcal{L}^\text{\RomanNumeralCaps{1}-L}_i = -2(\overline{\Delta}+\overline{\Phi}-1),
\end{equation}
which indicates this simple regularizer offers a gradient that relates to both error and evidence, enabling adaptive evidence suppression. To reinforce this constraint for the minority of extreme samples in a batch, we also introduce Type \RomanNumeralCaps{2}-line regularizers:
\begin{equation}
\label{loss:type2}
\mathcal{L}^\text{\RomanNumeralCaps{2}-L}_i(\boldsymbol{w})=\|\overline{\Phi}+\overline{\Delta}-1\|^{2}_2 - \|\overline{\Phi}-\overline{\Delta}\|^{2}_2,
\end{equation}
$\mathcal{L}^\text{\RomanNumeralCaps{2}-L}_i$ imposes constraints ensuring that the point $(\overline{\Delta}, \overline{\Phi})$ deviates from the line $\overline{\Delta} = \overline{\Phi}$, thereby exerting stricter control over extreme samples. In Appendix~\ref{grad}, we demonstrate that our regularizers effectively guide the optimization direction for evidence.

Our training objective for the evidential head is the combination of NLL and Geom-regularization:
\begin{equation}
\label{loss:der}
\mathcal{L}^\text{e}_i(\boldsymbol{w}) = 
\begin{cases} 
\lambda_\text{NLL}  \mathcal{L}^{\mathrm{NLL}}_{i} +  \lambda_\text{geom} \mathcal{L}^\text{I-L}_i(\boldsymbol{w}), & \text{Type \RomanNumeralCaps{1} } \\
\lambda_\text{NLL}  \mathcal{L}^{\mathrm{NLL}}_{i} +  \lambda_\text{geom}  \mathcal{L}^\text{II-L}_i(\boldsymbol{w}), & \text{Type \RomanNumeralCaps{2}}
\end{cases}
\end{equation}

To this end, our total loss can be formulated by a combination of grounding loss $\mathcal{L}_{G}$ (discussed in Appendix~\ref{VTG} ) and evidential loss:
\begin{equation}
\label{loss:all}
\mathcal{L} =\mathcal{L}_{G} + \lambda_\text{der} \frac{2}{N} \sum_{i=1}^N \mathcal{L}^\text{e}_i(\boldsymbol{w}),
\end{equation}
where $N$ symbolizes the number of clips in a training set.

\section{Experiment}
We focus on the following key considerations to conduct convincing experiments: \textbf{1)} Despite focusing on robustness and interpretability, does our proposed SRAM model demonstrate competitive performance relative to current state-of-the-art VTG models? \textbf{2)} Does the proposed Semantic Masking Alignment (SMA) and RFF blocks enhance performance in VTG tasks? \textbf{3)} Does SRAM give low uncertainty when performing high localization accuracy statistically, and vice versa? \textbf{4)} Is our proposed Geom-regularizer more robust than the vanilla regularizer (\emph{i.e.}, Eq.~\ref{vallina_reg})? \textbf{5)} Can the model output a high uncertainty score in various OOD scenarios to inform abnormality?


\begingroup
\setlength{\tabcolsep}{4pt} 
\renewcommand{\arraystretch}{1} 
\begin{table*}[t]
\caption{
Performance on QVHighlights with the backbone of \colorbox{yellow!10}{CLIP} or \colorbox{cyan!10}{CLIP and SlowFast}. \textbf{Bold} numbers indicate the best performance, and \underline{underlined} numbers indicate the second best performance. MR denotes Moment retrieval and HD denotes Highlight Detection.
}
\label{table_QVHighlight}
\vspace{0.1cm}
\centering
{\scriptsize 
\begin{tabular}{l|ccccccc|ccccccc}
\hlineB{2.5}
\multicolumn{1}{c|}{Split} & \multicolumn{7}{c|}{test split} & \multicolumn{7}{c}{val split} \\ \hline
\multicolumn{1}{c|}{\multirow{3}{*}{Method}} &  \multicolumn{5}{c}{MR}     & \multicolumn{2}{c|}{HD} & \multicolumn{5}{c}{MR}     & \multicolumn{2}{c}{HD} \\ \cline{2-15} 
\multicolumn{1}{c|}{} & \multicolumn{2}{c}{R1} & \multicolumn{3}{c}{MAP}  & \multicolumn{2}{c|}{\textgreater{}= Very Good} & \multicolumn{2}{c}{R1} & \multicolumn{3}{c}{MAP}  & \multicolumn{2}{c}{\textgreater{}= Very Good} \\ \cline{2-15} 
\multicolumn{1}{c|}{} & @0.5 & @0.7 & @0.5 & @0.75 & Avg. & MAP & HIT@1 & @0.5 & @0.7 & @0.5 & @0.75 & Avg. & MAP & HIT@1\\ \hlineB{2.5}
MCN~\cite{anne2017localizing} & 11.4 & 2.7 & 24.9 & 8.2 & 10.7& - & - & - & - & - & - & - & - & -\\
CAL~\cite{escorcia2019finding} & 25.5 & 11.5& 23.4& 7.7 & 9.9 & - & - & - & - & - & - & - & - & -\\
XML~\cite{lei2020tvr} & 41.8 & 30.4 & 44.6 & 31.7 & 32.1 & 34.5 & 55.3 & - & - & - & - & - & - & -\\
XML+\cite{lei2020tvr} & 46.7 & 33.5 & 47.9 & 34.7 & 34.9 & 35.4 & 55.1 & - & - & - & - & - & - & -\\ \hline
M-DETR~\cite{NEURIPS2021_62e09734}& 52.9 & 33.0 & 54.8 & 29.4 & 30.7 & 35.7 & 55.6 & 53.9 & 34.8 & - & - & 32.2 & 35.7 & 55.6 \\
UMT~\cite{Liu_2022_CVPR} & 56.2 & 41.2 & 53.4 & 37.0 & 36.1 & 38.2 & 60.0 & 60.3 & 44.3 & - & - & 38.6 & 39.9 & \underline{64.2} \\
UniVTG~\cite{lin2023univtg} & 58.9 & 40.9 & 57.6 & 35.6 & 35.5 & 38.2 & 61.0 & 59.7 & - & - & - & 36.1 & 38.8 & 61.8 \\ 
MomentDiff~\cite{li2024momentdiff} & 57.4 & 39.7 & 54.0 & 35.7 & 36.0 & - & - & - & - & - & - & - & - & - \\
EaTR~\cite{Jang_2023_ICCV} & - & - & - & - & - & - & - & 61.4 & 45.8 & 61.9 & \underline{41.9} & 41.7 & 37.2 & 58.7 \\
UnLoc-B~\cite{Yan_2023_ICCV} & - & - & - & - & - & - & - & 64.5 & \underline{48.8} & - & - & - & - & - \\
\hline
\rowcolor{yellow!10}
\textbf{SRAM}-\textit{Base}\footnotemark[1]  & - & - & - & - & - & - & - & 62.2 & 41.9 & 60.4 & 37.4 & 36.8 & 38.9 & 63.4 \\ 
\rowcolor{yellow!10}
\textbf{SRAM}-\textit{Large}\footnotemark[1] & - & - & - & - & - & - & - & 63.9 & 45.6 & 62.6 & 41.6 & 40.3 & 39.3 & 63.4 \\ 
\rowcolor{cyan!10}
\textbf{SRAM}-\textit{Base} & \underline{61.9} & 44.2 & \underline{61.0} & 39.4 & 38.4 & \underline{39.1} & \underline{62.9} & \textbf{65.4} & 47.2 & \textbf{63.9} & 41.1 & 40.5 & \textbf{40.2} & \textbf{64.3} \\ 
\rowcolor{cyan!10}
\textbf{SRAM}-\textit{Large} & \textbf{62.3} & \textbf{45.5} & \textbf{61.2} & \textbf{41.8} & \textbf{40.6} & \textbf{39.5} & \textbf{63.0} & \underline{65.0} & \textbf{49.4} & 63.4 & \textbf{44.0} & \textbf{43.0} & \underline{40.1} & \underline{63.4} \\ 
\hlineB{2.5}
\end{tabular}
}
\end{table*}
\endgroup

\begin{table*}[t]
    \centering
    \scriptsize
     \centering
    \setlength{\tabcolsep}{6pt} 
    \renewcommand{\arraystretch}{1} 
        \caption{Moment retrieval performances on TACoS, Charades-STA and ActivityNet Captions.}
    \label{table_mr}
     \centering
    \begin{tabular}{l|cccc|cccc|cccc}
    \hlineB{2.5}
    \multicolumn{1}{c|}{\multirow{2}{*}{Method}} & \multicolumn{4}{c|}{TACoS} & \multicolumn{4}{c|}{Charades-STA} & \multicolumn{4}{c}{Egoed-NLQ} \\ \cline{2-13} 
    \multicolumn{1}{c|}{}& R0.3 & R0.5 & R0.7 & mIoU & R0.3  & R0.5 & R0.7 & mIoU  & R0.3  & R0.5 & R0.7 & mIoU\\ \hline
    2D-TAN~\cite{2DTAN_2020_AAAI} & 40.0  & 28.0  & 12.9  & 27.2 & 58.8 & 46.0  & 27.5 & 41.3  & 4.3 & 1.8  & 0.6 & 3.4\\
    VSLNet~\cite{zhang2020span} & 35.5  & 23.5  & 13.2  & 25.0 & 60.3 & 42.7  & 24.1 & 41.6 & 4.5 & 2.4  & 1.0 & 3.5 \\ 
    M-DETR~\cite{NEURIPS2021_62e09734} & 38.0 & 24.7  & 12.0  & 25.5 & 65.8 & 52.1 & 30.6  & 45.5 & 4.34 & 1.8  & 0.65 & 3.5 \\
    QD-DETR~\cite{Moon_2023_CVPR}  & -   & -   & -   & - & -   & 57.3 & 32.6 & - & - & -  & - & - \\
    LLaViLo~\cite{ma2023llavilo} & - & - & - & - & - & 55.7 & 33.4 & -  & - & -  & - & -\\
    MomentDiff~\cite{li2024momentdiff} & 44.75 & 33.7  & -  & - & - & 55.6  & 32.4 & - & - & -  & - & - \\
    UniVTG~\cite{lin2023univtg} & \textbf{51.4} & \underline{35.0}  & \underline{17.4}  & \underline{33.6} & \underline{70.8} & \underline{58.0}  & \underline{35.7} & \underline{50.1} & \textbf{7.3} & \textbf{4.0}  & \underline{1.3} & \textbf{4.9} \\
    \hline
    \rowcolor{cyan!10}
    \textbf{SRAM}-\textit{Base} & \underline{50.2} & \textbf{37.3} & \textbf{19.4} & \textbf{33.9} & \textbf{71.6} & \textbf{60.2} & \textbf{38.0} & \textbf{51.6} & \underline{6.5} & \underline{3.5} & \textbf{1.4} & \underline{4.7} \\  
    \hlineB{2.5}
    \end{tabular}
    \vspace{-0.3cm}
\end{table*}

\begingroup
\setlength{\tabcolsep}{9pt} 
\renewcommand{\arraystretch}{1} 
\begin{table*}[t!]
    \caption{Video summarization results on TVsum. $\dagger$ denotes methods with audio modality. }
	\label{table_TVsum}
	\centering
	{\scriptsize
    \begin{tabular}{l|cccccccccc|c}
    \hlineB{2.5}
    \multicolumn{1}{c|}{Method} & VT        & VU        & GA        & MS        & PK        & PR        & FM        & BK   & BT   & DS   & Avg. \\ \hlineB{2.5}
    LIM-S~\cite{xiong2019less}  & 55.9      & 42.9      & 61.2      & 54.0      & 60.3      & 47.5      & 43.2      & 66.3 & 69.1 & 62.6 & 56.3 \\
    Trailer~\cite{wang2020learning} & 61.3      & 54.6      & 65.7      & 60.8      & 59.1      & 70.1      & 58.2      & 64.7 & 65.6 & 68.1 & 62.8 \\
    SL-Module~\cite{xu2021cross} & \underline{86.5}      & 68.7      & 74.9      & \textbf{86.2}    & 79.0      & 63.2      & 58.9      & 72.6 & 78.9 & 64.0 & 73.3 \\ 
    MINI-Net~\cite{hong2020mini}$\dagger$ & 80.6 & 68.3 & 78.2      & \underline{81.8}      & 78.1      & 65.8      & 57.8      & 75.0 & 80.2 & 65.5 & 73.2 \\
    UMT~\cite{liu2022umt}$\dagger$ & \textbf{87.5}      & 81.5      & 88.2      & 78.8      & 81.4      & \textbf{87.0} & \textbf{76.0} & 86.9 & \underline{84.4} & \textbf{79.6} & \underline{83.1} \\
    UniVTG~\cite{lin2023univtg} & 83.9 & \underline{85.1} & \underline{89.0} & 80.1 & \underline{84.6} & \underline{81.4} & 70.9 & \underline{91.7} & 73.5 & 69.3 & 81.0 \\ 
     \hline
     \rowcolor{cyan!10}
    \textbf{SRAM}-\textit{Base} & 85.4 & \textbf{93.0} & \textbf{92.5} & 81.4 & \textbf{87.2}& 79.6 & \underline{72.0} & \textbf{92.2} & \textbf{87.1} & \underline{75.3} & \textbf{84.6} \\ \hlineB{2.5}
    \end{tabular}
    }
    \vspace{-0.2cm}
\end{table*}
\endgroup

\subsection{Datasets and Implementation details}
\textbf{Datasets}. We conducted experiments on several widely used public datasets from diverse scenes: Charades-STA~\cite{gao2017tall} (in-door activities), QVHighlights~\cite{lei2021detecting} (untrimmed daily vlogs \& news), TACoS~\cite{regneri2013grounding} (cooking scenes), Ego4d-NLQ~\cite{grauman2022ego4d} (egocentric videos) and TVSum (YouTube videos). The detailed information including specific task domains and sizes for different datasets is reported in Appendix~\ref{DAIMP}~Table~\ref{tab:implementation} with their different hyperparameters.

\footnotetext[1]{We didn't report the performance on the test split for the limited number of submissions to the test server (to avoid overfitting). }
\textbf{Metrics}. For moment retrieval, we use recall@1 with IoU thresholds of 0.5 and 0.7, mean average precision (MAP) with IoU thresholds of 0.5 and 0.75, and MAP avg, which is the average MAP across IoU thresholds from 0.5 to 0.95 in 0.05 increments.  For Highlight detection and Video summarization, we use MAP. Following~\cite{NEURIPS2021_62e09734}, an additional metric HIT@1 is utilized for the Highlight detection task in the QVHighlights dataset, representing the hit ratio of the highest-scored clip.

\textbf{Experimental Settings}. Following previous works~\cite{lin2023univtg,li2024momentdiff}, we utilize CLIP~\cite{radford2021learning} (ViT-B/32) and SlowFast~\cite{feichtenhofer2019slowfast} (ResNet-50)  as a frozen backbone. 
We conducted experiments with two model sizes, referred to as "base" and "large," by setting the hidden dimension in the RFF Blocks to 512 and 1024, respectively. Unless otherwise specified, the number ($\mathbf{n}$) of RFF blocks is set to 4. The training process is divided into two stages. In the first stage, SMA masks and reconstructs noun entities in the query. Each sentence has 1 noun masked by default. The default epoch for SMA is set to 30, with a learning rate of 1e-5. We utilize spaCy’s transformer-based parser-~\cite{spacy2} to extract noun entities from the query text, and the masking is accomplished by zeroing out the noun entities at the embedding level. In the second stage, SRAM predicts bounding boxed on the visual branch at each video clip. Since our purpose of using DER is to optimize uncertainty without affecting the model's grounding capability, the gradient of delta in Eq.~\ref{loss:type1} and~\ref{loss:type2} is set to zero. To avoid similar predictions, we utilize NMS with a threshold of 0.7 at evaluation to achieve better performance. If not stated otherwise, we used the Type \RomanNumeralCaps{1}-line regularizer on the evidential head. All training for the moment retrieval tasks are conducted on four Tesla V100 GPUs. For the video summarization task, due to the smaller scale of the TVSum dataset, we used only a single V100 GPU.
\vspace{-0.2cm}
\subsection{Quantitative results}

\textbf{Comparison with the state-of-the-art.} To demonstrate the effectiveness of SRAM, we compared it with 17 state-of-the-art methods across different datasets. As reported in Table~\ref{table_QVHighlight}, SRAM outperforms existing methods on various metrics. 
On the QVHighlights test set, for the Moment Retrieval task, SRAM-Large obtains 62.26\% for R1@0.5 and 45.53\% for R1@0.7. For the Highlight Detection task, SRAM-Large reaches 63.04\% for HIT@1. In the Moment Retrieval task, SRAM-Large outperforms MomentDiff~\cite{li2024momentdiff} by an average of 5.72\%, and in the Highlight Detection task, it surpasses UniVTG~\cite{lin2023univtg} by an average of 1.65\%.
According to Table \ref{table_mr}, we supplement the performance comparison on the SRAM-Base model for the TACoS, Charades-STA, and Ego4d-NLQ datasets. On the TACoS dataset, SRAM-Base outperforms the nearest competitor UniVTG by 2.3\% and 2.0\% respectively. For the Charades-STA dataset, SRAM-Base surpasses previous results by 0.8\%, 2.2\%, and 2.3\% respectively. We observe that the Eg4d-NLQ dataset has distinct characteristics such as long video durations (8-20 minutes) and high textual ambiguity (mainly consisting of questions, \emph{i.e.}, "When did I put my shirt into the closet?"). These features pose significant challenges for accurate grounding moments.
To validate SRAM's performance in video summarization, we present a comparison of the TVSum dataset in Table~\ref{table_TVsum}. For each domain in TVSum, SRAM has generally demonstrated strong performance. Specifically, in the VU domain, SRAM outperforms UniVTG by 8\%. Additionally, on the overall average metrics, SRAM exceeds UniVTG by 3.6\% and surpasses the UMT model, which utilizes audio modality, by 1.5\%.

\begin{wraptable}{r}{0.5\textwidth}
\centering
\vspace{-0.4cm}
\captionof{table}{Ablation study of RFF block, F denotes whether to use flipped cross-attention and n denotes the number of RFF blocks.}
\begin{tabular}{>{\centering\arraybackslash}p{0.5cm}cccccc}
\hlineB{2.5}
F & n & R1@0.5 & R1@0.7 \\
\hline
\ding{55} & 2 & 54.58 & 32.58 \\
\ding{51} & 2 & 56.13 & 33.81 \\
\ding{51} & 4 & 58.94 & 40.39 \\
\ding{51} & 6 & 58.19 & 39.55 \\
\hlineB{2.5}
\label{tab:RFF_block}
\end{tabular}
\vspace{-1cm}
\end{wraptable}
\textbf{Ablation study.} We first validated the effectiveness of SMA pretraining. Figure~\ref{fig:ablation_mlm} (a) shows the model performance differences under different settings of epochs. Notably, when the number of SMA epochs increased from 0 to 50, there was a significant improvement in MAP (+1.5\%).  Furthermore, we tested the effectiveness of the proposed RFF blocks. As shown in Table 4, keeping the number of RFF blocks constant, using flipped cross attention significantly improved R1@0.5 by 1.55\% compared to using cross attention separately in the visual and textual branches (split cross attention). A deeper ablation analysis of DER with our proposed Geom-regularizer is reported in Appendix~\ref{sec:geom_appendix}.

\textbf{Parameter analysis.} We analyze the impact of SMA epochs, SMA learning rate, weights of DER, and Geom-regularizer on the model's grounding performance. The results can be seen in Appendix~\ref{subsec:analysis}.

\subsection{Qualitive resluts}

\textbf{Uncertainty Calibration.}
\label{scatter}
Since we propose geom-regularization to enhance the calibration of model predictions, we aim to assess the efficacy of our approach by contrasting the performance of aleatoric and epistemic uncertainty estimation with and without our regularization technique, as well as against using vanilla regularization in\cite{amini2020deep}. Ideally, optimal uncertainty measures should effectively identify deviations in predictions (\emph{i.e.}, take high uncertainty when the model is making errors). Figure~\ref{fig:calibration} in Appendix~\ref{u_carlibration} illustrates our comparison of different regularization methods on the QVHighlights val set. The horizontal axis of each scatter plot represents $\overline{\Delta}$ (\emph{i.e.} normalized error), while the vertical axis represents one of the two types of uncertainty.



\begin{figure}[h]
\vspace{-0.3cm}
\centering
\includegraphics[width=0.8\linewidth]{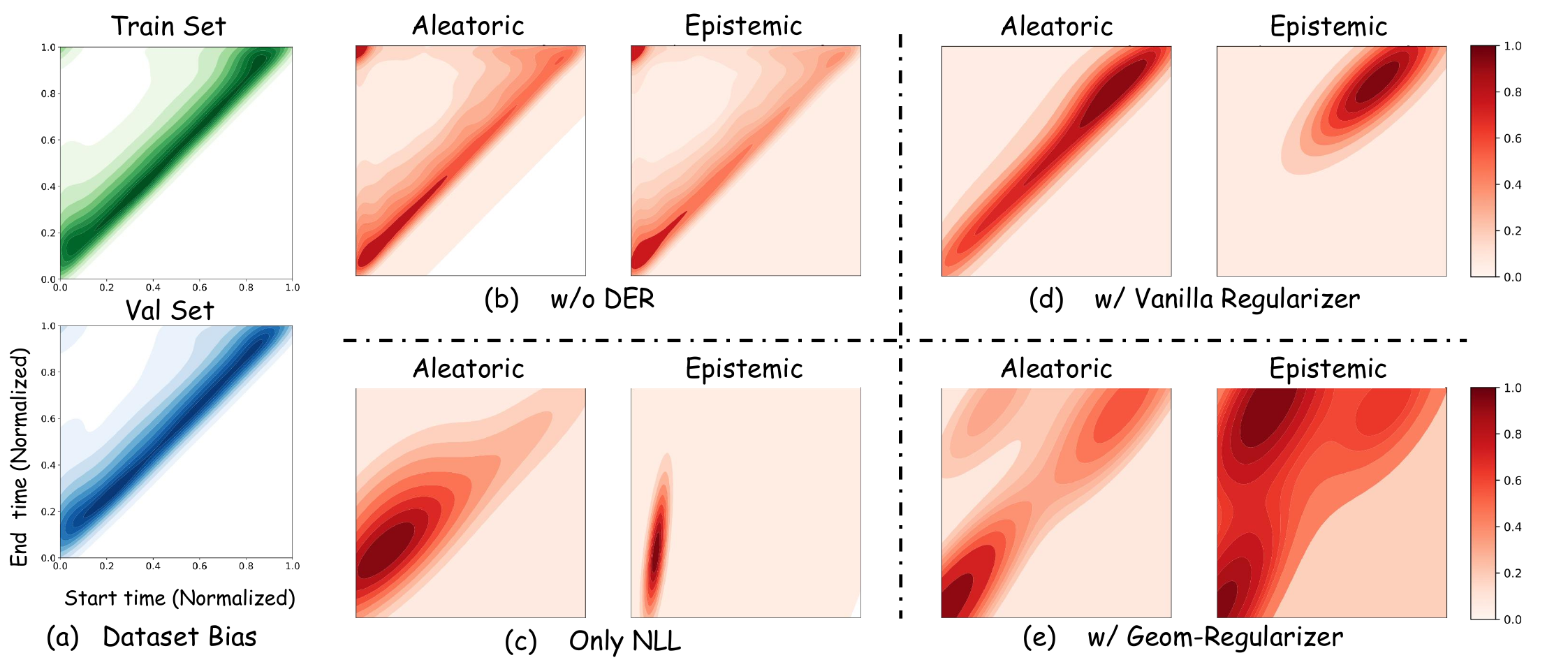}
\vspace{-0.1cm}
\caption{\textbf{Dataset bias sensitivity.} (a) Joint distributions of the start and end timestamps of the ground-truth moments in the QVHighlights dataset. (b), (c), (d), and (e) show the predicted uncertainty's sensitivity to temporal biases in the dataset under different conditions.}
\label{fig:bias_qv}
\end{figure}

\textbf{Bias sensitivity.} As previously reported in studies, most Moment Retrieval datasets exhibit significant imbalances in the duration and position of moments. As shown in Figure~\ref{fig:bias_qv} (a), using the QVHighlights dataset as an example, we visualize the joint distribution of the normalized start times and end times of all ground truth moments. The light-colored areas in the figure indicate regions with almost no moment distribution, leading to \textit{\textbf{Temporal OOD}}. Higher epistemic uncertainty is demanded when samples belong to the Temporal OOD region. 
We analyzed the uncertainty predicted by the model in different time regions under various experimental settings in the QVHighlights dataset.
Figure~\ref{fig:bias_qv} (b) shows that without DER constraints, the evidential head's predicted aleatoric and epistemic uncertainty tends to simply fit the biased temporal distribution.
Figure~\ref{fig:bias_qv} (c) shows that using only the NLL constraint (\emph{i.e.}, Eq.~\ref{NLL_Loss}), most regions exhibit extremely low epistemic uncertainty, indicating the model is overly confident in its predictions.
Figure~\ref{fig:bias_qv} (d) illustrates that with Vanilla regularization, although this approach does suppress the concentration of uncertainty in a specific region, it does not show sensitivity to OOD data.
Figure~\ref{fig:bias_qv} (e) demonstrates that with our proposed Geom-regularization term, the temporal OOD regions exhibit significantly higher epistemic uncertainty. 

\textbf{Adversarial experiments}
To further validate the robustness of SRAM in OOD scenarios, we conducted adversarial experiments using the QVHighlights validation set. Specifically, we gradually add noise to (a) video embeddings (\emph{i.e.}, Figure~\ref{fig:vid_noise}), (b) text embeddings, as shown in Figure~\ref{fig:txt_noise}, and (c) both text and video embeddings (\emph{i.e.}, Figure~\ref{fig:all_noise}). Using Gaussian kernel density estimation (KDE), we plotted the uncertainty distribution of the predictions for the entire validation set. It can be observed that as the level of noise increases, the model's epistemic uncertainty gradually shifts from smaller to larger values. Notably, when noise is added to both modalities simultaneously, the uncertainty exhibits a more pronounced increase.

\textbf{Case study.} We constructed a typical adversarial example to demonstrate the effectiveness of SRAM. For example, if the video depicts "\textit{A \textbf{plane} is flying in the sky}," the model outputs very low epistemic uncertainty when given the correct textual query. However, when given an adversarial query (\emph{e.g.} "\textit{A \textbf{bird} is flying in the sky}"), the misalignment between the visual content and the semantics causes SRAM to exhibit higher aleatoric and epistemic uncertainty. More cases can be seen in Appendix~\ref{fig:cases_study}.

\section{Conclusion}
As the development of Artificial General Intelligence (AGI) progresses, increasingly sophisticated VTG models are emerging. However, these models often falter when confronted with open-ended user inputs. Addressing this challenge, this paper introduces a robust VTG model, namely SRAM, that not only possesses VTG capabilities but also enables explicit and comprehensive quantification of potential uncertainties. This allows the model to provide credible responses to queries that exceed its operational scope, paving the way for future reliable, video-driven HCIs. Limited by data quality and scale, the model's modality alignment capabilities are not particularly notable. Nevertheless, it offers strategies for enhancing the trustworthiness of AI decisions. Future research will focus on further expanding the decision-making reliability and interpretability of multimodal large language models in video-related downstream tasks.

\newpage

\bibliographystyle{plainnat} 
\bibliography{neurips_2024}



\appendix
\newpage
\section{Derivations}
\label{sec:sup-derivations}

\subsection{Normal Inverse-Gamma moments}
\label{sec:sup-nig-derivations}

We assume our data was drawn from a Gaussian with unknown mean and variance, $(\mu, \sigma^2)$. We probabilistically model these parameters, $\bm\theta$, according to: 
\begin{align}
    \mu \sim \mathcal{N}(\gamma, \sigma^2 \upsilon^{-1}) \\
    \sigma^2 \sim \Gamma^{-1}(\alpha, \beta).
\end{align}
Therefore, the prior joint distribution can be written as: 
\begin{align}
    p(\underbrace{\mu, \sigma^2}_{\bm\theta} | \underbrace{\gamma, \upsilon, \alpha, \beta}_{\bm \varphi }) &= p(\mu) \, p(\sigma^2)\\
    &= \mathcal{N}(\gamma, \sigma^2 \upsilon^{-1}) \, \Gamma^{-1}(\alpha, \beta)\\
    &= \frac{\beta^{\alpha}\sqrt{\upsilon}}{\Gamma(\alpha)\sqrt{2 \pi \sigma^{2}}} \left(\frac{1}{\sigma^{2}}\right)^{\alpha+1} \exp \left\{-\frac{2 \beta+\upsilon(\gamma-\mu)^{2}}{2 \sigma^{2}}\right\}.
\end{align}

The first-order moments of this distribution represent the maximum likelihood prediction as well as uncertainty (both aleatoric and epistemic).

\begin{align}
    \E[\mu] &= \int^\infty_{\mu=-\infty} \mu \, p(\mu) \diff\mu = \gamma
\end{align}
\begin{align}
    \E[\sigma^2] &= \int_{\sigma^2=0}^\infty \sigma^2 \, p(\sigma^2) \diff\sigma^2  \\
    &= \int_{\sigma=0}^\infty\sigma^2 \, p(\sigma^2) \, (2\sigma) \diff\sigma\\
    &= \frac{\beta}{\alpha-1}, \qquad \forall\, \alpha > 1
\end{align}
\begin{align}
    \Var[\mu] &= \int_{\mu=-\infty}^\infty \mu^2 \, p(\mu) \diff\mu - (\E[\mu])^2  \\
    &= \gamma^2 - \frac{\sigma^2}{\upsilon} - (\E[\mu])^2  \\
    &= \gamma^2 - \frac{\frac{\beta}{\alpha-1}}{\upsilon} - \gamma^2  \\
    &= \frac{\beta}{\upsilon(\alpha-1)}, \qquad \forall\, \alpha > 1.
\end{align}
In summary, 
\begin{align}
    \underbrace{\E[\mu]=\gamma}_{\text{prediction}}, \qquad \underbrace{\E[\sigma^2]=\tfrac{\beta}{\alpha-1}}_{\text{aleatoric}}, \qquad \underbrace{\Var[\mu]=\tfrac{\beta}{\upsilon(\alpha-1)}}_{\text{epistemic}}.
\end{align}
\label{eq:sup-nig-moments}

\subsection{Model evidence \& Type II Maximum Likelihood Loss}
\label{sec:sup-loss-derivations}
In this subsection, we derive the posterior predictive or model evidence (\emph{i.e.} Eq.~\ref{eq:post_pred}) of a \nig{} distribution. Marginalizing out $\mu$ and $\sigma$ gives our desired result: 
\begin{equation}
p(b_i|\bm \varphi ) = \text{St}\left(b_i; \gamma, \frac{\beta(1+ \upsilon)}{\upsilon\,\alpha} , 2\alpha\right).
\label{eq:post_pred}
\end{equation}

\begin{align}
p(b_i | \bm \varphi ) & = \int_{\bm \theta} p(b_i|\bm\theta) p(\bm\theta | \bm \varphi ) \, \diff\bm\theta \\
& = \int_{\sigma^2=0}^{\infty} \int_{\mu=-\infty}^{\infty} p(b_i|\mu, \sigma^2) p(\mu, \sigma^2 | \bm \varphi ) \, \diff\mu \diff\sigma^2 \\
& = \int_{\sigma^2=0}^{\infty} \int_{\mu=-\infty}^{\infty} p(b_i|\mu, \sigma^2) p(\mu, \sigma^2 | \gamma, \upsilon, \alpha, \beta) \, \diff\mu \diff\sigma^2 \\
& = 
\int_{\sigma^2=0}^{\infty} \int_{\mu=-\infty}^{\infty}
\left[
             \sqrt{\frac{1}{2 \pi \sigma^2}} \exp \left\{-\frac{(b_i-\mu)^{2}}{2\sigma^2}\right\}
\right] \\
& \quad \quad
\left[
 \frac{\beta^{\alpha}\sqrt{\upsilon}}{\Gamma(\alpha)\sqrt{2 \pi \sigma^{2}}} \left(\frac{1}{\sigma^{2}}\right)^{\alpha+1} \exp \left\{-\frac{2 \beta+\upsilon(\gamma-\mu)^{2}}{2 \sigma^{2}}\right\}
 \right]
 \, \diff\mu \diff\sigma^2 
 \\
& = \int_{\sigma^2=0}^{\infty}
\frac{\beta^\alpha \sigma^{-3-2\alpha}}{\sqrt{2\pi} \sqrt{1+1/\upsilon} \Gamma(\alpha)}
\exp \left\{ - \frac{2\beta + \frac{\upsilon (b_i - \gamma)^2}{1+\upsilon}}{2\sigma^2}\right\}
\diff\sigma ^2 
\\
& = \int_{\sigma=0}^{\infty}
\frac{\beta^\alpha \sigma^{-3-2\alpha}}{\sqrt{2\pi} \sqrt{1+1/\upsilon} \Gamma(\alpha)}
\exp \left\{ - \frac{2\beta + \frac{\upsilon (b_i - \gamma)^2}{1+\upsilon}}{2\sigma^2}\right\}
2\sigma \diff\sigma 
\\
& = \frac{\Gamma(1/2 + \alpha)}{\Gamma(\alpha) }  \sqrt{\frac{\upsilon}{\pi}}
\left(2\beta (1+\upsilon)\right)^\alpha
\left( \upsilon (b_i-\gamma)^2 + 2\beta (1+\upsilon)\right)^{-(\frac{1}{2}+\alpha)} \\
p(b_i | \bm \varphi ) &= \text{St}\left(b_i; \gamma, \frac{\beta(1+ \upsilon)}{\upsilon\,\alpha} , 2\alpha\right).
\end{align}

$\text{St}\left(\bm b; \mu_\text{St}, \sigma_\text{St}^2, \upsilon_{St}\right)$ is the Student-t distribution evaluated at $\bm b$ with location parameter $\mu_\text{St}$, scale parameter $\sigma_\text{St}^2$, and $\upsilon_{\text{St}}$ degrees of freedom. Using this result we can compute the negative log-likelihood loss, $\Lnll_i$, for sample $i$ as: 
\begin{align}
\Lnll_i &= -\log p(b_i|\bm \varphi ) \\
&= -\log \left( \text{St}\left(b_i; \gamma, \frac{\beta(1+ \upsilon)}{\upsilon\,\alpha} , 2\alpha\right) \right) \\
\Lnll_i &= \tfrac{1}{2}\log\left(\tfrac{\pi}{\upsilon}\right) - \alpha\log(\Omega) + \left(\alpha+\tfrac{1}{2}\right) \log((b_i-\gamma)^2\upsilon + \Omega) + \log\left( \tfrac{\Gamma(\alpha)}{\Gamma(\alpha+\frac{1}{2})}\right),
\end{align}

where $\Omega = 2\beta(1+\upsilon)$. 

\section{Datasets And Implementation Details.}
\subsection{Parameters of datasets}
\label{DAIMP}
In Table~\ref{tab:implementation}, we list the datasets used in this study, including dataset size, task category, video clip length, and detailed hyperparameters used for model training.

\begin{table}[h]
    \centering
    \scriptsize
    \setlength{\tabcolsep}{3pt} 
    \renewcommand{\arraystretch}{1} 
    \caption{VTG dtasets list. \textbf{MR} denotes Moment Retrieval, \textbf{HD} denotes Highlight Detection, and \textbf{VS} denotes Video Summarization. \textbf{S} means seconds, \textbf{LR} denotes learning rate, \textbf{Epochs} denotes total training epochs, \textbf{Warm-up} means number of warm-up iterations, and \textbf{LR Drop} means the epoch that drops learning rate by $1/10$.}
    \begin{tabular}{l|ccc|c|cccccccc}
    \hlineB{2.5}
    \textbf{Dataset} & \textbf{MR} & \textbf{HD} & \textbf{VS}  & \textbf{\string# Samples} &  \textbf{S} & \textbf{Batch Size} & \textbf{LR} & \textbf{Epochs} & \textbf{Warm-up} & \textbf{LR Drop} & \textbf{SMA LR} & \textbf{SMA Epoch} \\
    \hline
    QVHighlights \cite{lei2021detecting} & \checkmark & \checkmark &  & 10.3K & $2$ & $32$ & $1e^{-4}$ & $200$ & $10$ & $180$ & $1e^{-5}$  & $30$\\
     Charades-STA \cite{gao2017tall} & \checkmark &  & & 16.1K &$1$ & $32$ & $1e^{-4}$ & $100$ & $10$ & $50$ & $1e^{-5}$  & $10$\\
    TACoS \cite{regneri2013grounding} & \checkmark &  & & 18.2K & $2$ & $32$ & $1e^{-4}$ & $150$ & $10$ & $80$  & $1e^{-5}$  & $30$\\
     Ego4d-NLQ \cite{grauman2022ego4d} & \checkmark &  & & 15.1K & $2$ & $64$ & $1e^{-5}$ & $200$ & $10$ & $20$  & $1e^{-5}$  & $30$\\
     TVSum \cite{song2015tvsum} &  & \checkmark &  & 50 & $2$ & $4$ & $1e^{-3}$ & $400$ & $50$ & N/A  & $1e^{-5}$  & $10$\\
    \hlineB{2.5}
    \end{tabular}
    \label{tab:implementation}
\end{table}

\subsection{Implementations for normalizations}
\label{norm}
\textbf{Normalization:}
We have tried two normalization operations, \emph{i.e.}min-max normalization and using activation function to normalize. 
\begin{itemize}
\item 
\textbf{Min-Max normalization:}
Assume we have evaluated an increasing sequence of errors, that is:
\begin{equation}
\{\Delta_1, \Delta_2, \cdots, \Delta_n\}
\end{equation}
where $n$ represents batch size. Min-Max Normalization maps $\Delta_i$ to $\overline{\Delta_i}$ by:
\begin{equation}
\overline{\Delta_i} = \frac{\Delta_i}{\Delta_n - \Delta_1}
\end{equation}
We recommend using this normalization method in training and batch testing.
\item 
\textbf{Normalization using activation functions:}
Use activation functions $tanh(\cdot)$ so that we can map $\Delta_i$ to $\overline{\Delta_i}$, which is between 0 and 1:
\begin{equation}
\overline{\Delta_i} = tanh(\Delta_i)
\end{equation}
And $\Phi$ is normalized in the same way to $\overline{\Phi}$. 
We recommend this normalization for single-point or small-batch testing.
\end{itemize}


\textbf{Histogram equalization:}
Although we normalize the uncertainty, we still find that the distribution of uncertainty is extremely biased to 0. 
We consider that this is still due to the overconfidence effect that NLL brings to the model. In order to obtain a more expressive uncertainty estimation in the inference process, sometimes we use histogram equalization to post-process the normalized uncertainty.

\begin{algorithm}[H]
    \renewcommand{\algorithmicrequire}{\textbf{Input:}}
    \renewcommand{\algorithmicensure}{\textbf{Output:}}
    \caption{Histogram Equalization}
    \label{alg:hist_eq}
    \begin{algorithmic}[1] 
        \REQUIRE Sequence of values: $X = \{\Delta_1, \Delta_2, \ldots, \Delta_n\}$
        \ENSURE Equalized sequence: $X' = \{\overline{\Delta_1}, \overline{\Delta_2}, \ldots, \overline{\Delta_n}\}$
        \STATE $hist \leftarrow \text{calculate\_histogram}(X)$
        \STATE $cdf \leftarrow \text{calculate\_cdf}(hist)$
        \STATE $X' \leftarrow \text{apply\_cdf\_mapping}(X, cdf)$
        \RETURN $X'$
    \end{algorithmic}
\end{algorithm}

\section{Qualitative analysis}
\subsection{Visulization of attention map}
\begin{figure}[h]
\centering
\includegraphics[width=\linewidth]{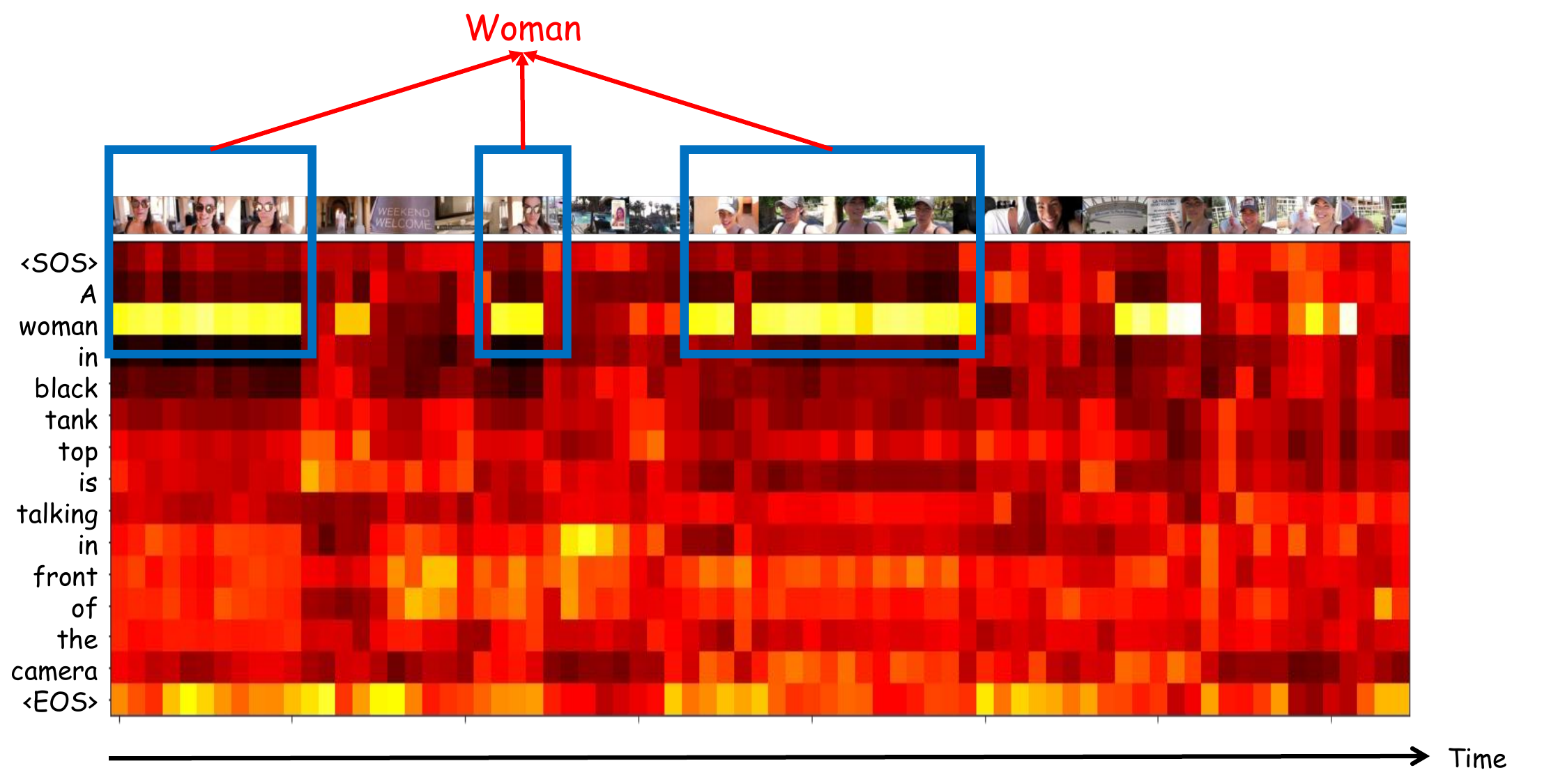}
\vspace{-0.5cm}
\caption{\textbf{Case \RomanNumeralCaps{1} of attention map visualization.} }
\label{fig:att_case1}
\end{figure}

\begin{figure}[h]
\centering
\includegraphics[width=\linewidth]{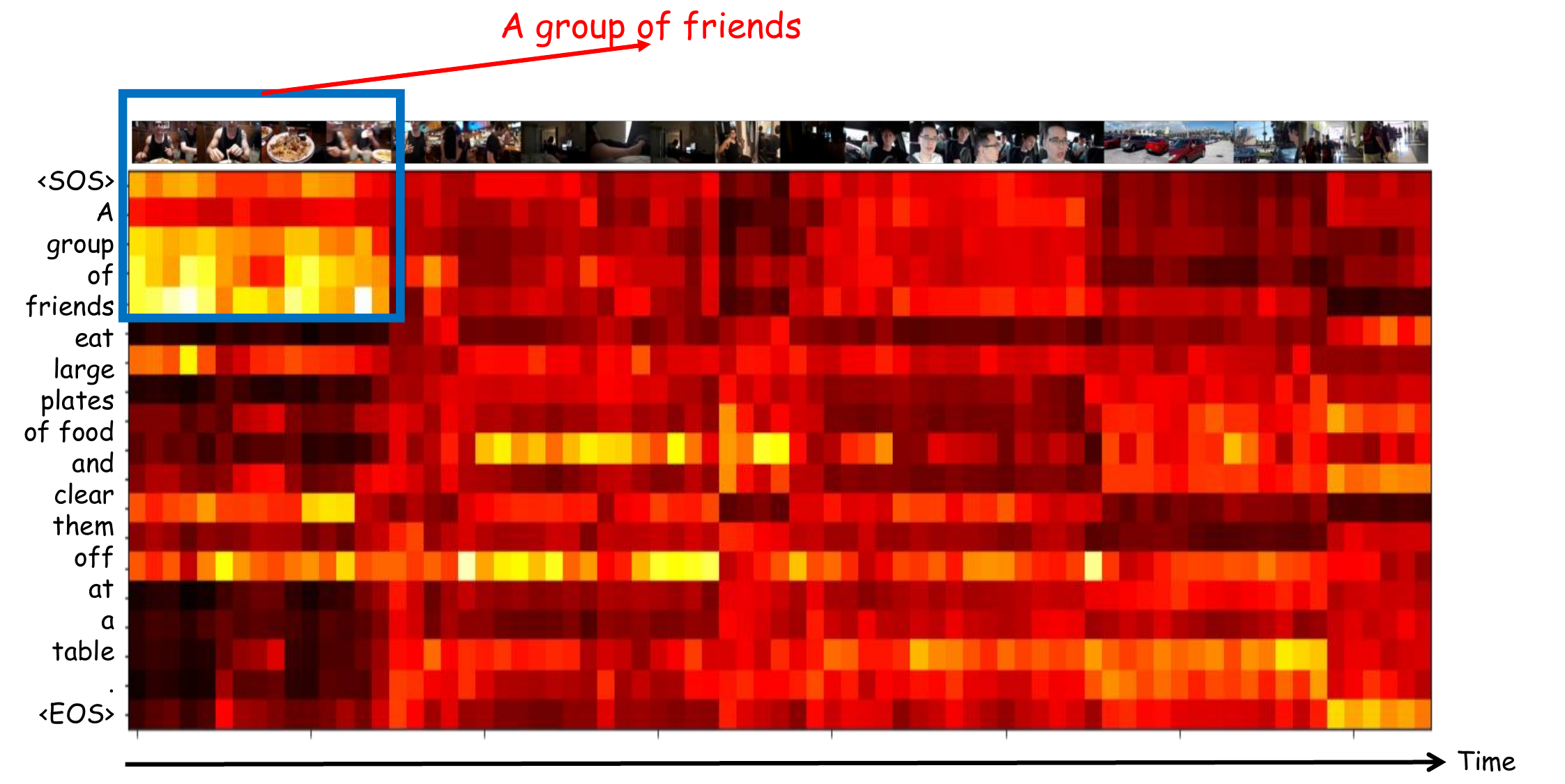}
\vspace{-0.5cm}
\caption{\textbf{Case \RomanNumeralCaps{2} of attention map visualization.} }
\label{fig:att_case2}
\end{figure}

The visualizations provided in Figure~\ref{fig:att_case1} and Figure~\ref{fig:att_case2} demonstrate the effectiveness of our SRAM in achieving fine-grained cross-modal alignment in VTG. Figure~\ref{fig:att_case1} effectively maps the attention to the visual cues of a woman speaking, aligning precisely with the textual description, thereby enhancing video-to-text translation accuracy. Similarly, Figure~\ref{fig:att_case2} shows that SRAM is capable to focus on a group of friends interacting around a table, accurately reflecting the descriptive text, which is essential for generating contextually accurate video summaries. These examples underscore the potential of SRAM not only to improve downstream task performance by ensuring temporal and contextual relevance but also to serve as a basis for investigating the model's uncertainty. By analyzing where the model allocates attention, researchers can identify areas of high confidence and potential uncertainty, aiding in the refinement of VTG models for more reliable and transparent AI-driven applications.

\newpage
\subsection{Visualization of uncertainty calibration}
\label{u_carlibration}
\begin{figure*}[h]
\vspace{-0.3cm}
\centering
\includegraphics[width=\linewidth]{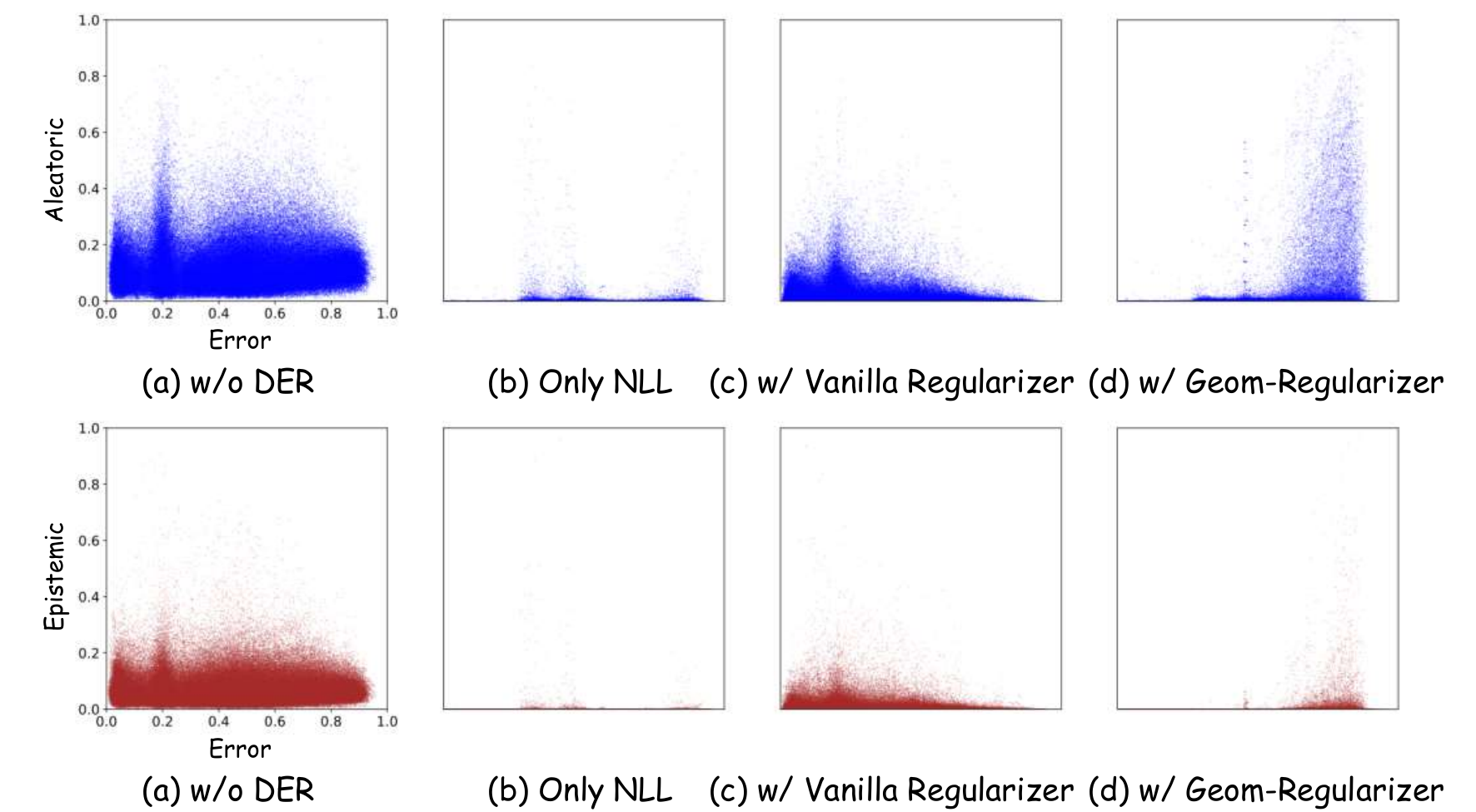}
\vspace{-0.1cm}
\caption{Effects of Various Regularization Techniques on Aleatoric and Epistemic Uncertainty Distribution. Panels (a)-(d) illustrate the impact of different regularization methods on the relationship between aleatoric uncertainty (top row) and epistemic uncertainty (bottom row) with respect to prediction error. The models include: (a) without DER, (b) only NLL, (c) with Vanilla Regularizer, and (d) with Geom-Regularizer. Furthermore, it enriches information of uncertainty prediction compared to the one with only NLL as demonstrated in Table~\ref{tab:aleatoric_uncertainty} and Table~\ref{tab:epistemic_uncertainty}.}
\label{fig:calibration}
\end{figure*}
Figure~\ref{fig:calibration} demonstrates the influence of different regularization strategies on model uncertainty in relation to prediction error. The top row focuses on aleatoric uncertainty, which is inherent data uncertainty, whereas the bottom row examines epistemic uncertainty, which stems from model ignorance. And we can discern the following key information:

\begin{itemize}
    \item \textbf{(a) Without DER}: This model lacks any form of uncertainty management in the absence of DER, leading to inference results that are difficult to trust due to the complete absence of handling latent uncertainties.

    \item \textbf{(b) Only NLL}: In this configuration, the model exhibits extremely low uncertainty across all levels of error rates, indicating overconfidence due to overfitting. This overconfidence suggests a model that is not realistically cautious about its predictions.

    \item \textbf{(c) With Vanilla Regularizer}: Although the vanilla regularizer in DER measures and manages uncertainty, it paradoxically induces the model to express higher uncertainty at lower error rates and very low uncertainty at higher error rates. This counterintuitive behavior is clearly problematic, as it does not align with rational expectations of uncertainty behavior.

    \item \textbf{(d) With Geom-Regularizer}: Compared to (a), our proposed Geom-regularizer effectively measures and manages uncertainty, enabling the model to indicate higher uncertainty at higher error rates and vice versa. Relative to (b), it successfully mitigates the model's overconfidence, which is beneficial for making prudent decisions. Against (c), it accurately calibrates the measurement of uncertainty, achieving a more sensible and intuitive assessment of uncertainty levels.
\end{itemize}

\subsection{Visulization of gradient field}
\label{grad}
\begin{figure}[h]
\centering
\includegraphics[width=\linewidth]{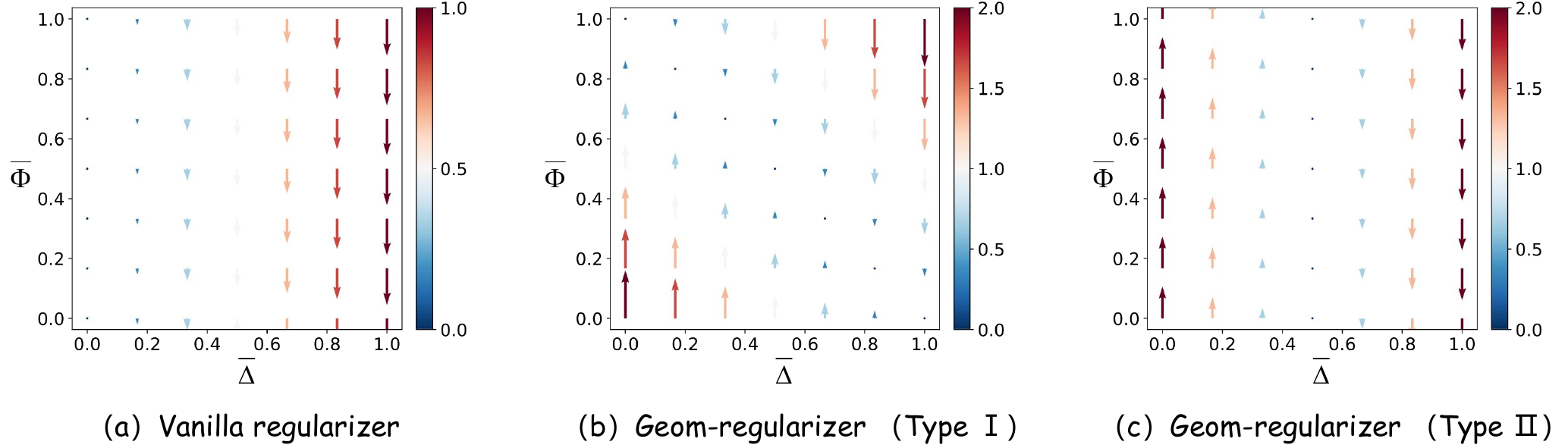}
\vspace{-0.5cm}
\caption{\textbf{Gradient field visualizations for regularizers}: (a) Vanilla regularizer used in DER~\cite{amini2020deep}, (b) Type \RomanNumeralCaps{1} Geom-regularizer, (c) Type \RomanNumeralCaps{2} Geom-regularizer.}
\label{fig:grad}
\end{figure}
Figure~\ref{fig:grad} illustrates the impact of different regularization strategies on the evidence (\emph{i.e.} $\overline{\Phi}$) adjustments in DER. Panel (a) reveals that the vanilla regularizer applies penalties based solely on error magnitude, with a tendency to decrease evidence as error increases. This approach often results in insufficient gradients for batches dominated by small errors, potentially leading to biased uncertainty assessments. Conversely, panels (b) and (c) showcase Type \RomanNumeralCaps{1} and Type \RomanNumeralCaps{2} Geom-regularizers, which modulate penalties dynamically based on both error magnitude and current evidence levels. This adaptive penalization facilitates more accurate evidence adjustments across varying error scenarios, thus addressing the limitations observed with the vanilla regularizer. Type \RomanNumeralCaps{2} regularizer extends this principle by applying aggressive penalties near critical points $(0,1), (1,0)$, enhancing the model's capability to consistently refine evidence in response to error-evidence dynamics, ultimately improving uncertainty quantification in DER models.

\subsection{Cases study}
\begin{figure*}
\vspace{-0.3cm}
\centering
\includegraphics[width=\linewidth]{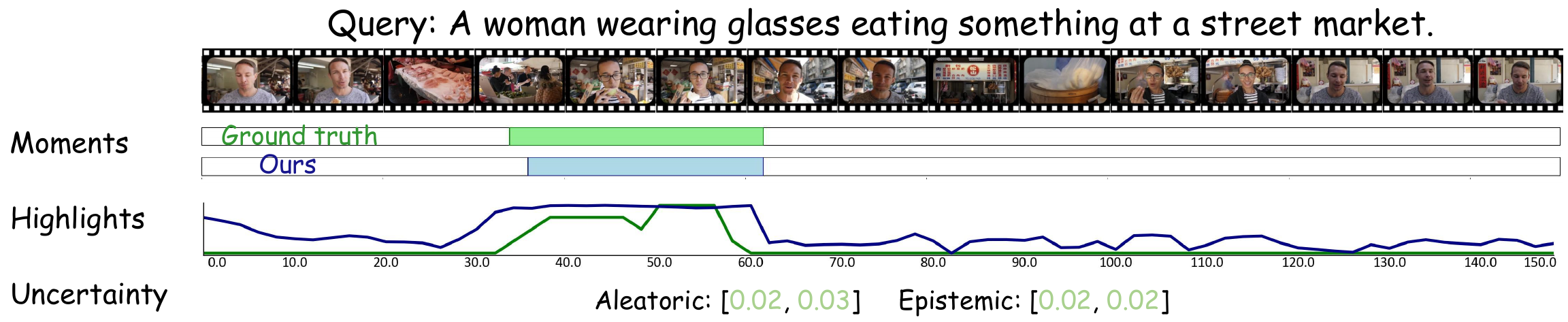}
\includegraphics[width=\linewidth]{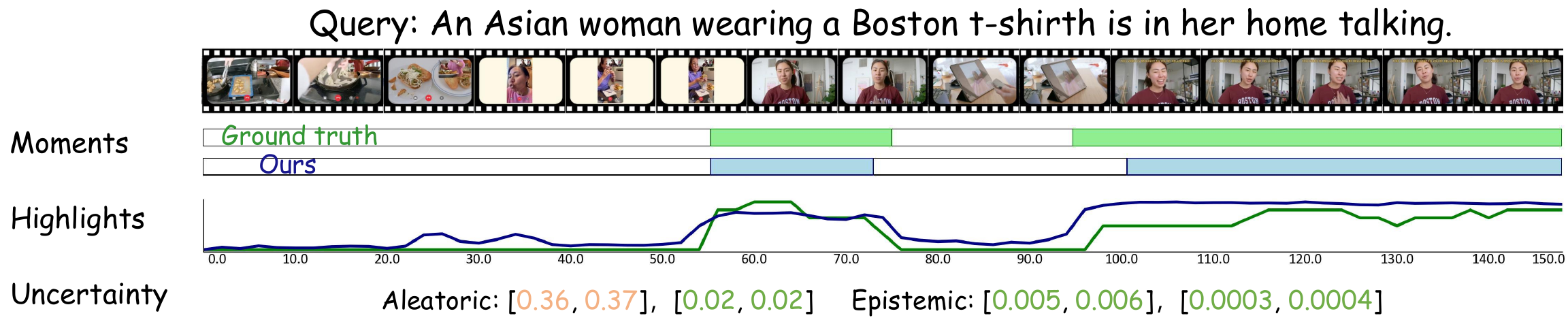}
\includegraphics[width=\linewidth]{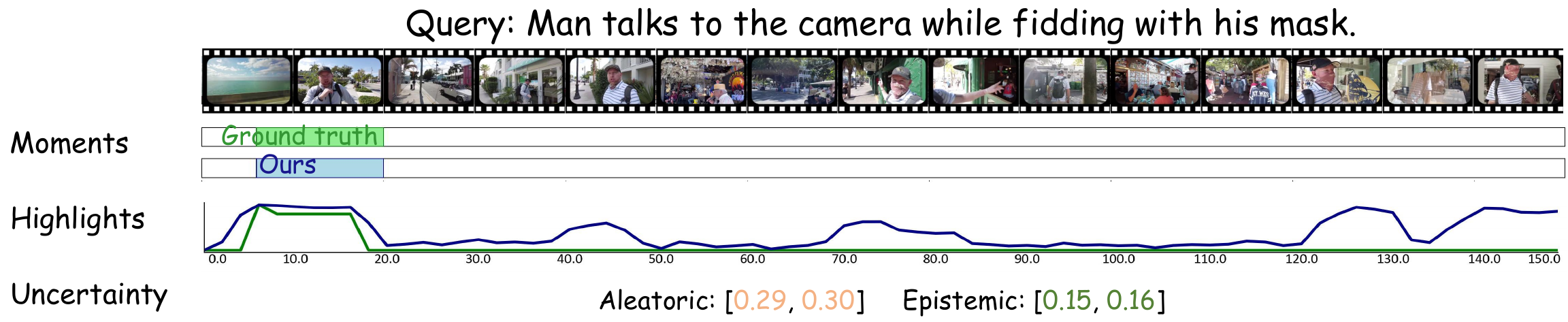}
\includegraphics[width=\linewidth]{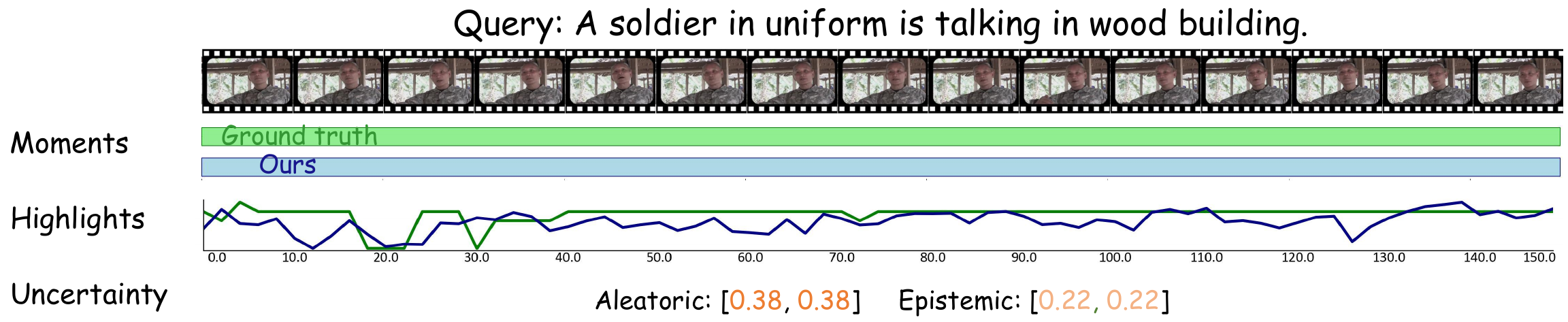}
\includegraphics[width=\linewidth]{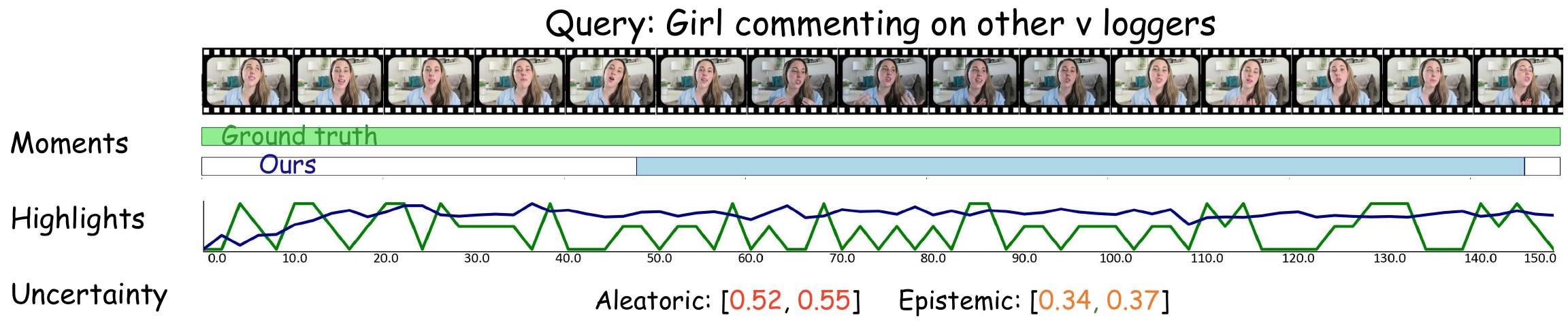}
\includegraphics[width=\linewidth]{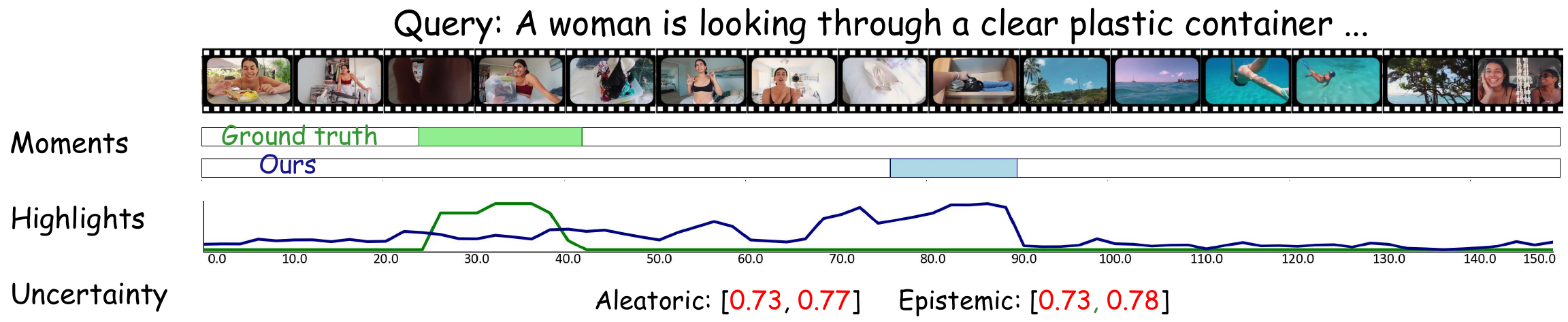}
\includegraphics[width=\linewidth]{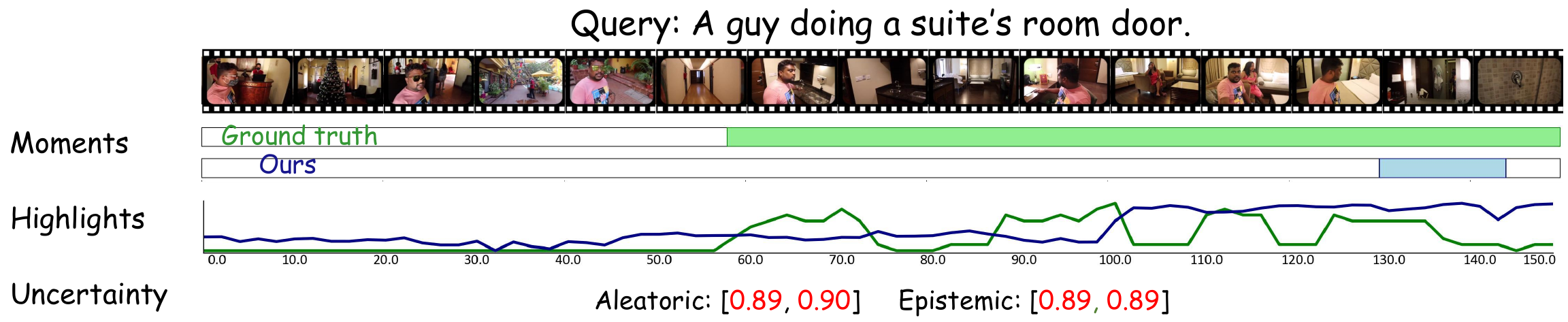}
\caption{\textbf{Cases Study}}
\label{fig:cases_study}
\end{figure*}

In figure~\ref{fig:cases_study}, we select some cases from the validation set of QVHighlights that support the effectiveness of our model. For example, we can easily observe that highly accurate predictions are often accompanied with very low uncertainty, while highly inaccurate predictions are accompanied with very high uncertainty, as shown by the first case and the last two cases. Additionally, when there exist scene changes (case 2, case 3) or changes in lighting conditions (case 5) in the video, the model is also prone to output higher uncertainty, especially aleatoric uncertainty.
\label{cases_study}

\subsection{Visulization of error-evidence evolution}
\label{sec:evo}
\begin{figure*}
\vspace{-0.3cm}
\centering
\includegraphics[width=\linewidth]{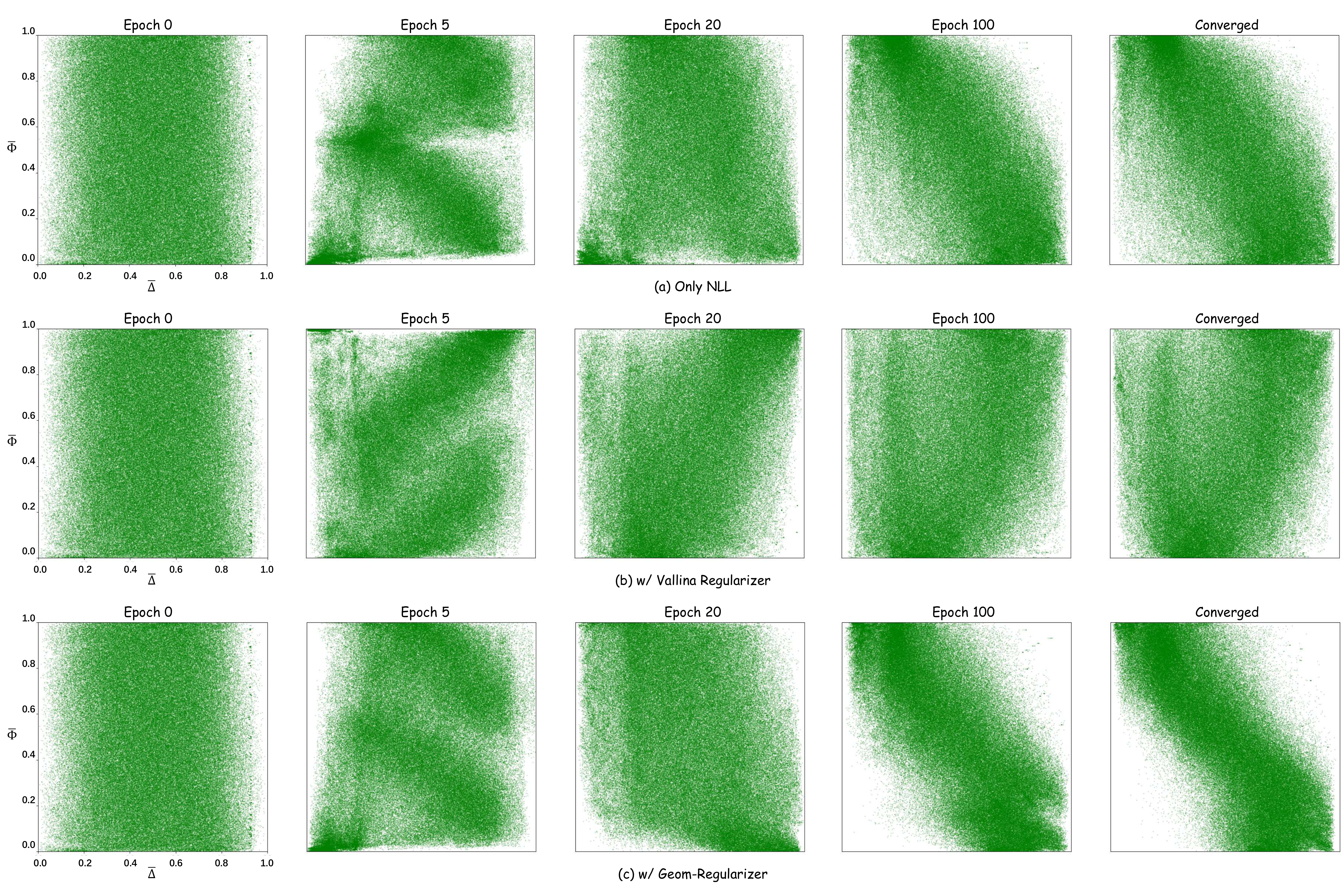}
\vspace{-0.3cm}
\caption{\textbf{Evolution of the predicted \((\mathbf{\mathbf{\overline{\Delta}},\overline{\Phi}})\)s' distribution over training epochs with different regularization techniques on  QVHighlights~\cite{lei2021detecting}.} This figure showcases how the evidence \((\overline{\Phi})\) and error \((\overline{\Delta})\) distributions evolve across training epochs (0, 5, 20, 100, and convergence) under three regularization strategies: (a) only NLL, (b) added vanilla regularizer, and (c) our Geom-regularizer. 
}
\label{fig:evo}
\end{figure*}
As illustrated in Figure~\ref{fig:evo}, it is obvious that "\textbf{\textit{accurate predictions with high evidence while inaccurate predictions with low evidence}}" has been reflected in the knowledge of model with only NLL. Unfortunately, the vanilla regularizer excessively suppress the evidence of low error predictions, but ignores and even enlarges the evidence of high error predictions. Geom-regularizer turn the situation around, retain the main knowledge learned by NLL, and provides calibration for more reasonable uncertainty estimation.

\subsection{Adversarial experiments}
\label{ad_noise}
We conduct adversarial experiments on SRAM at the statistical level and case level, in order to demonstrate that SRAM really capture increased predictive uncertainty on samples that have been adversarily perturbed.

\begin{figure*}
\vspace{-0.3cm}
\centering
\includegraphics[width=\linewidth]{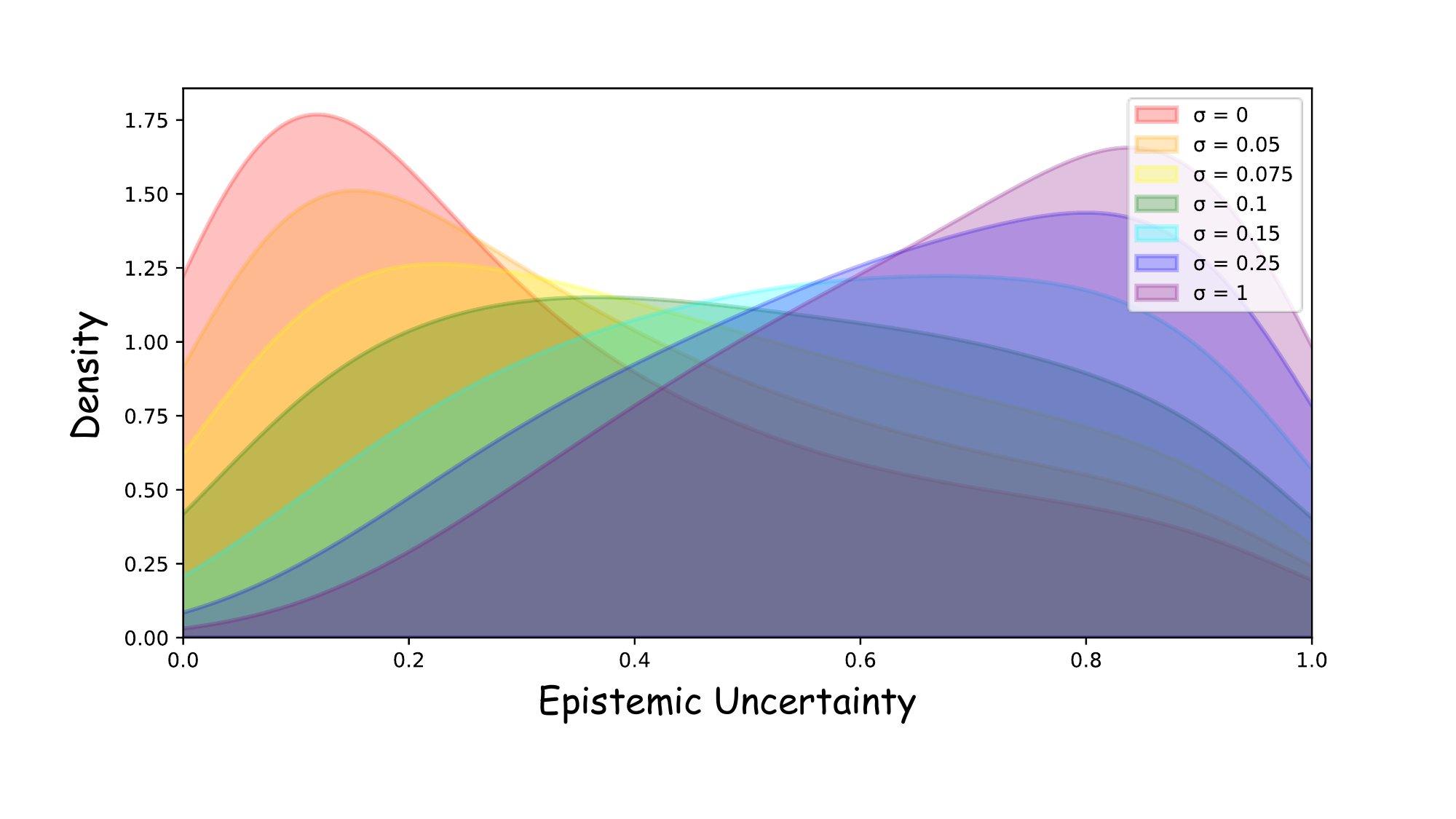}
\vspace{-1cm}
\caption{\textbf{Adversarial Experiments.} We add Gaussian noise of different variances to video features and compare their distribution of epistemic uncertainty.}
\label{fig:vid_noise}
\end{figure*}

\begin{figure*}
\vspace{-0.3cm}
\centering
\includegraphics[width=\linewidth]{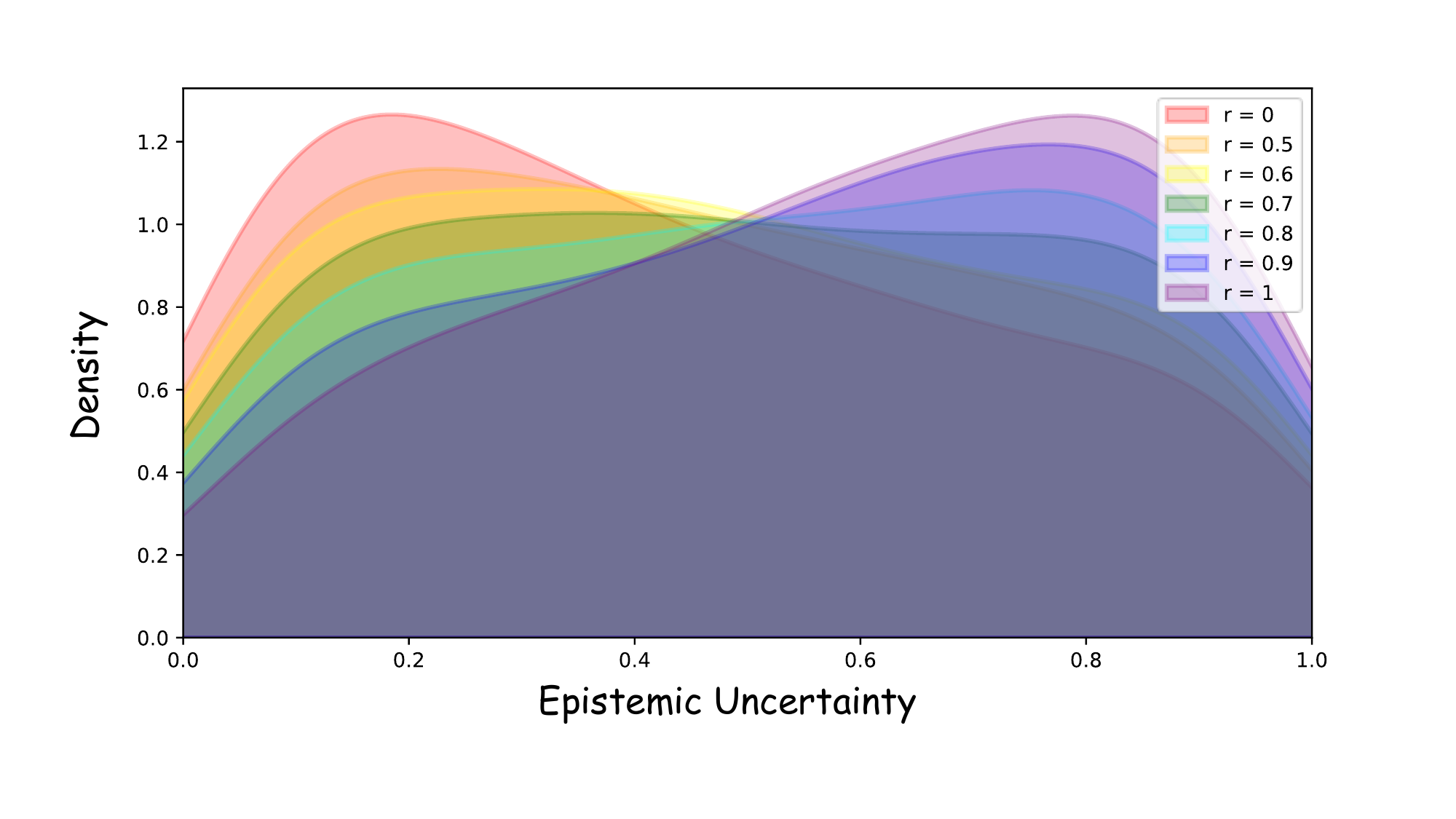}
\vspace{-1cm}
\caption{\textbf{Adversarial Experiments.} We add noise to the text by replacing tokens with tokens of other sequences and compare the distribution of epistemic uncertainty.}
\label{fig:txt_noise}
\end{figure*}

\begin{figure*}
\vspace{-0.3cm}
\centering
\includegraphics[width=\linewidth]{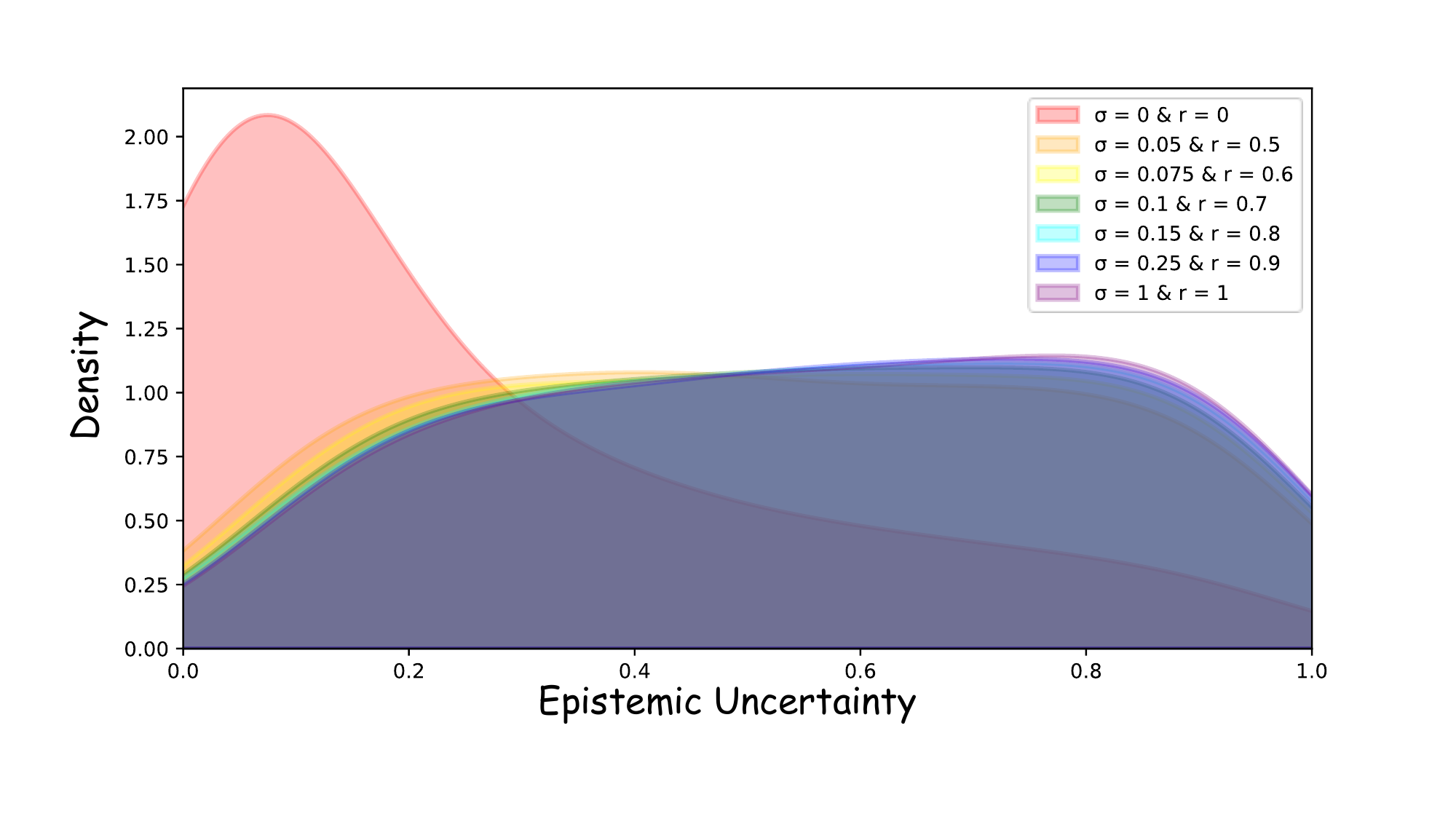}
\vspace{-1cm}
\caption{\textbf{Adversarial Experiments.} We use the same noise schedule as uni-modality situations and add noise to both modalities at the same time. 
It can be seen that even with the same noise intensity, adding noise to both modalities simultaneously will greatly enlarge the epistemic uncertainty.}
\label{fig:all_noise}
\end{figure*}

For the statistical level, we conduct three experiments on the validation set of QVHighlights: add noise only to the visual modality, only to the text modality and to both modalities. For visual modality, we add Gaussian noise of different variances to video features. As shown in Figure~\ref{fig:vid_noise}, with the increase of noise intensity, the center of distribution does change from 0 to 1. For text modality, we uniformly replace tokens of a certain proportion $r$ in each text sequence with the token in the same position of other sequences in a batch, and ensure that new tokens in different positions come from different sequences. As shown in Figure~\ref{fig:txt_noise}, with the increase of replacement proportion $r$, the distribution moves in the desired direction. As for adding noise to both modalities at the same time, we set the variances of Gaussian noise and replacement proportions $r$ to be consistent with single-modality experiments. In this case, uncertainty increases much faster than that in the uni-modality situation, as shown in Figure~\ref{fig:all_noise}, which further proves the ability of SRAM to estimate uncertainty.  

\begin{figure}[h]
\vspace{-0.3cm}
\centering
\includegraphics[width=\linewidth]{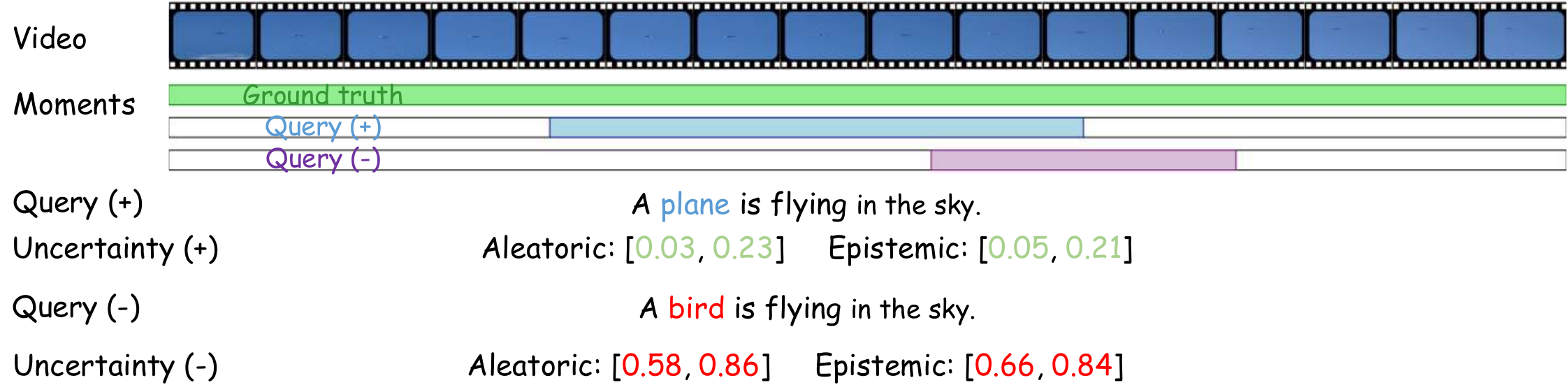}
\vspace{-0.1cm}
\caption{\textbf{Adversarial Case \RomanNumeralCaps{1}.} We select a semantically ambiguous video, where the plane is extremely small, making it difficult for even humans to discern whether it is an airplane or a bird. We provide both the correct query (query with "plane") and an incorrect query (replacing "plane" with "bird") to SRAM, compare the uncertainty of their highest confidence predictions, and find that the model assign higher uncertainty to the incorrect query.}
\label{fig:plane}
\end{figure}

\begin{figure}[h]
\vspace{-0.3cm}
\centering
\includegraphics[width=\linewidth]{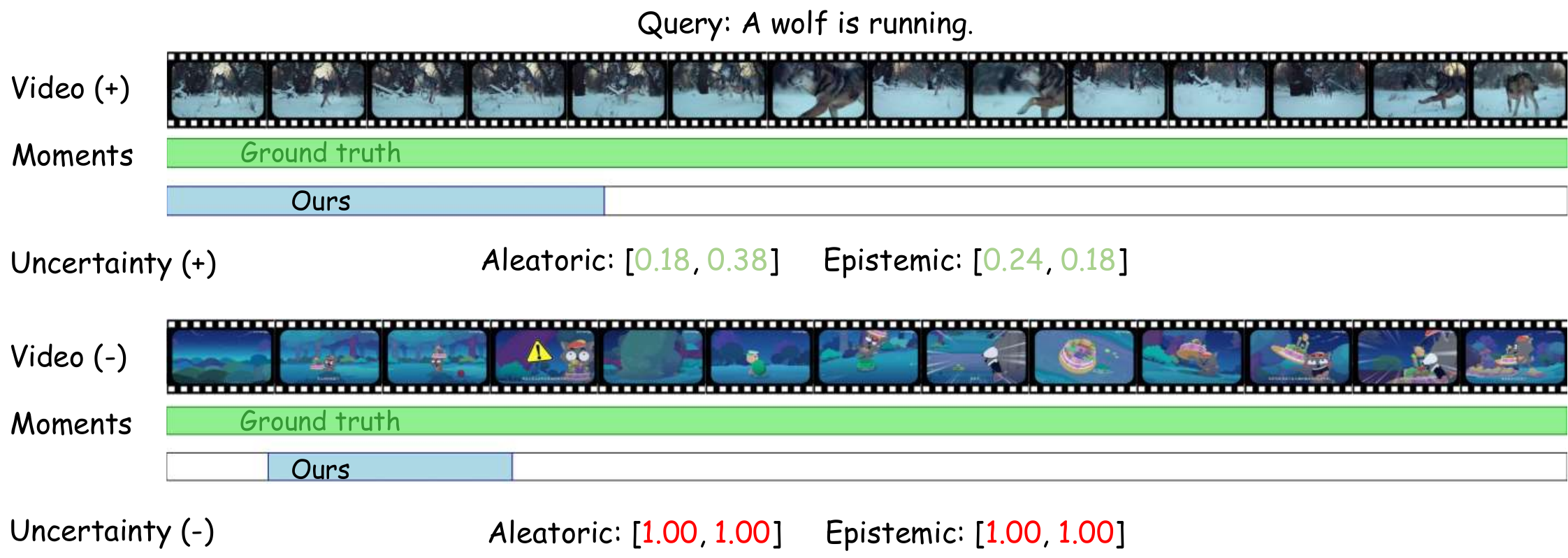}
\vspace{-0.1cm}
\caption{\textbf{Adversarial Case \RomanNumeralCaps{2}.} We select a real video and an animated video of "a running wolf" and provide the model with the same query "a wolf is running". It can be observed that SRAM outputs higher certainty for the animated video.}
\label{fig:wolf}
\end{figure}

\begin{figure}[h]
\vspace{-0.3cm}
\centering
\includegraphics[width=\linewidth]{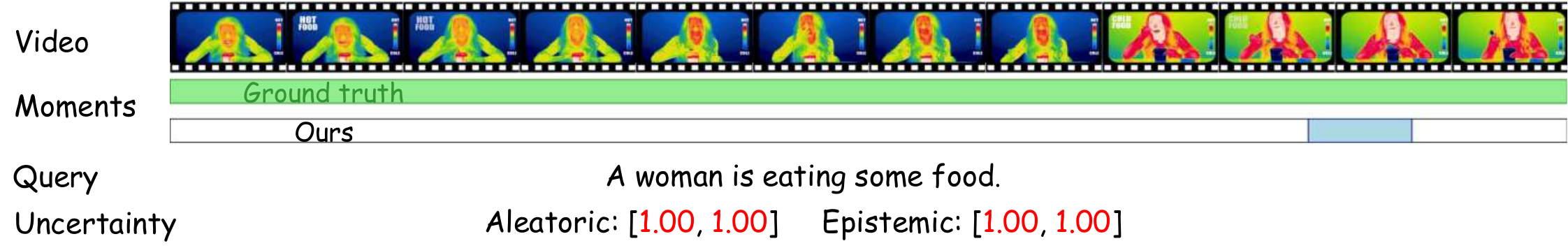}
\vspace{-0.1cm}
\caption{\textbf{Adversarial Case \RomanNumeralCaps{3}.} We select an infrared thermal imaging video, a type that the model has rarely encountered in the training set. As an OOD video, SRAM assign it very high uncertainty.}
\label{fig:red}
\end{figure}

\begin{figure}[h]
\vspace{-0.3cm}
\centering
\includegraphics[width=\linewidth]{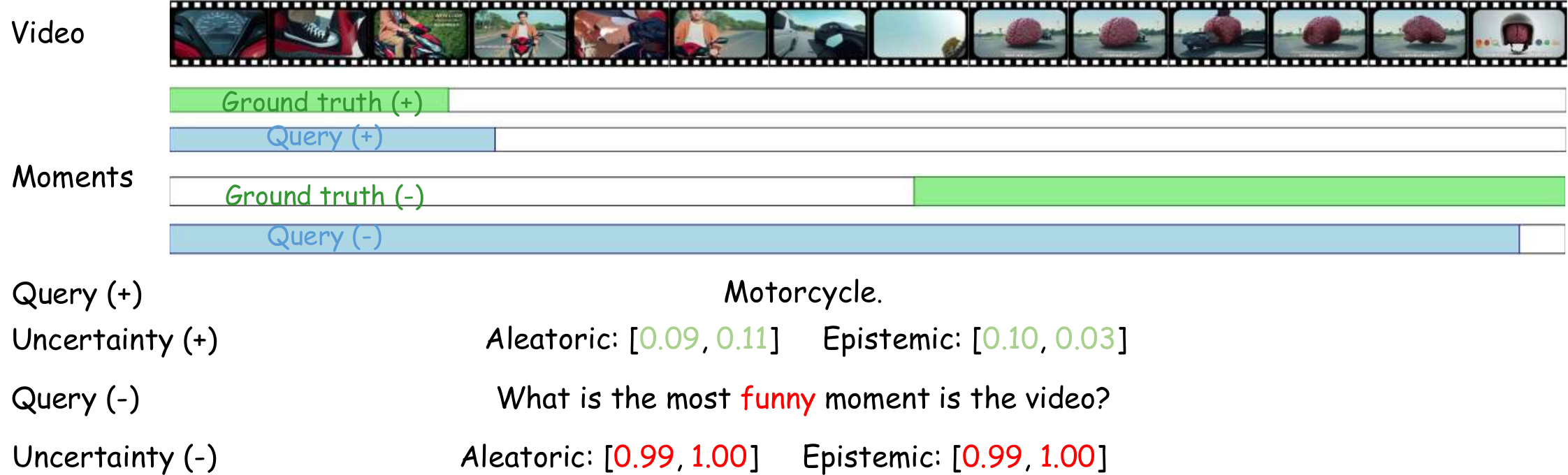}
\vspace{-0.1cm}
\caption{\textbf{Adversarial Case \RomanNumeralCaps{4}.} We select an advertisement video containing humor. We provide SRAM with both simple query and abstract query for prediction. Our results showed that SRAM struggle to provide accurate localization for the abstract queries but exhibit high uncertainty.}
\label{fig:funny}
\end{figure}

For the case level, we design four experiments to show SRAM is able to capture a high degree of uncertainty in some specific situations that is very likely to exist in reality, which has been discussed in \label{motivation}. In Figure~\ref{fig:plane}, 
SRAM demonstrate effective perception of ambiguous visual semantics, providing predictions while also outputting higher uncertainty. In Figure~\ref{fig:wolf}, SRAM assign higher uncertainty to the OOD video (which is a cartoon) , even though both videos contain the semantic "a wolf is running". In Figure~\ref{fig:red}, SRAM also assign higher uncertainty to the OOD video, which is infrared thermal imaging video. In Figure~\ref{fig:funny}, the word "funny" in query is abstract and confuses the model, but SRAM successfully provides high uncertainty to compensated for the failure in prediction.

\newpage
\section{Quantitative analysis}
\subsection{Quantitative analysis of Geom-regularizer}
\label{sec:geom_appendix}
To measure the degree to which the uncertainty predicted under different regularization settings obeys the prior of \textbf{"the larger the error, the greater the uncertainty"}, we define measure: Error-Uncertainty Consistency Measure (EUCM). EUCM is calculated as: 
\begin{equation}
EUCM=\|\overline{\Delta}+\overline{\mathcal{U}}\|^2_2,
\end{equation}
where $\mathcal{U}$ represents uncertainty. 
Moreover, we also compute the information entropy of different uncertainty distributions, which is used to evaluate the expressive ability of evidential predictor.
We tested EUCM and entropy on the validation set of QVHighlights~\cite{lei2021detecting} in Table~\ref{tab:aleatoric_uncertainty} and~\ref{tab:epistemic_uncertainty} under different settings.
\begin{table*}[t!]
\caption{Ablation studies: Aleatoric uncertainty metrics on QVHighlights.}
\label{tab:aleatoric_uncertainty}
\centering
\renewcommand{\arraystretch}{1.2}
\begin{tabular*}{\textwidth}{@{\extracolsep{\fill}}l|cccc@{}}
\hlineB{2.5}
\multicolumn{1}{c|}{Metric} & NLL & w/ Vanilla-Reg. & w/ 1-line & w/ 2-line \\
\hlineB{2.5}
EUCM $\downarrow$ & 0.3144 & 0.3045 & \textbf{0.2815} & 0.2835 \\
Entropy $\uparrow$ & 0.1775 & \textbf{0.3561} & 0.2713 & 0.2916 \\
\hlineB{2.5}
\end{tabular*}
\vspace{-0.2cm}
\end{table*}
\begin{table*}[t!]
\caption{Ablation studies: Epistemic uncertainty metrics on QVHighlights.}
\label{tab:epistemic_uncertainty}
\centering
\renewcommand{\arraystretch}{1.2} 
\begin{tabular*}{\textwidth}{@{\extracolsep{\fill}}l|cccc@{}}
\hlineB{2.5}
\multicolumn{1}{c|}{Metric} & NLL & w/ Vanilla-Reg. & w/ 1-line & w/ 2-line \\
\hlineB{2.5}
EUCM $\downarrow$ & 0.3147 & 0.3104 & 0.3030 & \textbf{0.2964} \\
Entropy $\uparrow$ & 0.0114 & \textbf{0.3536} & 0.3516 & 0.2230 \\
\hlineB{2.5}
\end{tabular*}
\vspace{-0.2cm}
\end{table*}
We can notice that the entropy of predictions with the vanilla regularizer is greater than that with the Geom-regularizer. However, as demonstrated in section~\ref{scatter} predictions made with the vanilla regularizer are prone to be misleading. Consequently, even though it exhibits higher information entropy, this entropy may encompass substantial "wrong information". In contrast, the Geom-regularizer not only achieves higher information entropy but also results in a lower EUCM score, indicating its superior performance.

\subsection{Parameter analysis}
\label{subsec:analysis}
\begin{figure}[h]
\centering
\includegraphics[width=\linewidth]{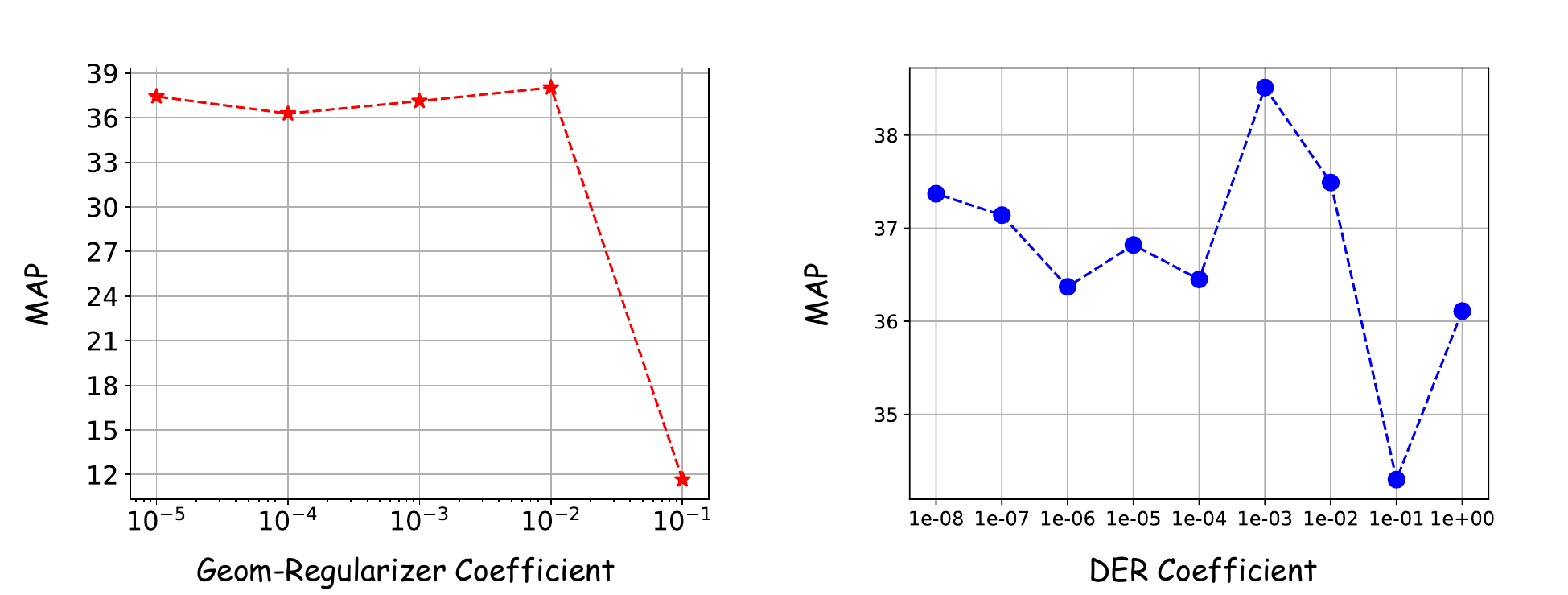}
\caption{Parameters Analysis of DER.{We examined the change of MAP. The left one results under different coefficients for Geom-regularizer, and the right one results under different coefficients for DER.}}
\label{fig:der}
\vspace{-0.1cm}
\label{fig:para_der}
\end{figure}

\textbf{Parameters analysis of DER.}
As shown in Figure~\ref{fig:der}, we examined the change in MAP as \(\lambda_{\text{der}}\) gradually increased from \(1 \times 10^{-8}\) to \(1 \times 10^{0}\). We observed that when \(\lambda_{\text{der}}\) is small, the model's performance remains unaffected. However, as \(\lambda_{\text{der}}\) increases to \(1 \times 10^{-2}\), MAP begins to decline, reaching its lowest at \(1 \times 10^{-1}\). This can be explained by the fact that an excessively high uncertainty constraint weight forces the model's optimization direction to overfit the evidential head rather than maintaining its basic grounding ability. Therefore, we set \(\lambda_{\text{der}}\) to \(1 \times 10^{-3}\).

\begin{figure}[h]
\centering
\includegraphics[width=\linewidth]{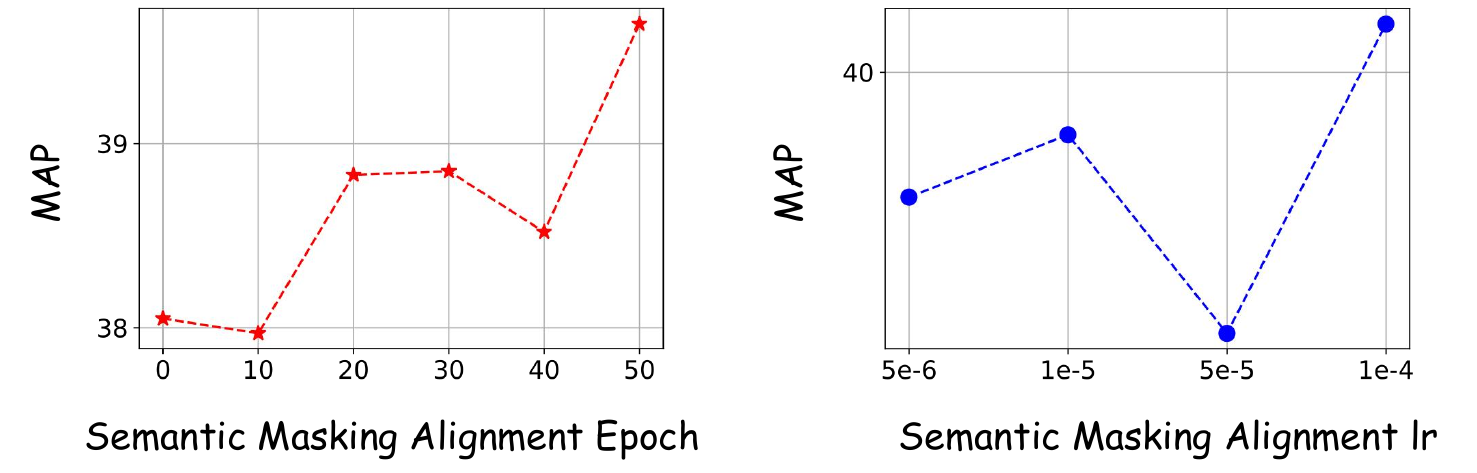}
\caption{Effectiveness of the SMA. We examined the change of, with (a) the number of SMA epochs or (b) the learning rate increasing gradually.}
\vspace{-0.1cm}
\label{fig:ablation_mlm}
\end{figure}

\textbf{Parameters analysis of SMA.}
Figure~\ref{fig:ablation_mlm} (b) shows the model performance differences under different learning rates in SMA.  We ultimately set the SMA learning rate to 1e-5 to achieve a smooth optimization. Furthermore, we tested the effectiveness of the proposed RFF blocks. As shown in Table~\ref{tab:RFF_block}, keeping the number of RFF blocks constant, using flipped cross attention significantly improved R1@0.5 by 1.55\% compared to using cross attention separately in the visual and textual branches (split cross attention). We perform a parameter analysis illustrated in the Figure~\ref{fig:para_der} (\emph{i.e.} see Appendix~\ref{sec:geom_appendix}).

\section{Detailed Workflow of the RFF Block}
\label{sec:workflow}
The RFF block processes inputs from the video and text branches, with initial features denoted as $V^{(1)}$ and $Q^{(1)}$ respectively. The cross-attention module, which shares parameters, alternates the roles of the video and text branches as queries and keys/values:
\begin{align}
  CA_{v \to q}^{(1)} &= \text{Softmax}\left(\frac{V^{(1)} Q^{(1)\mathsf{T}}}{\sqrt{d_k}}\right) Q^{(1)} \\
  CA_{q \to v}^{(1)} &= \text{Softmax}\left(\frac{Q^{(1)} V^{(1)\mathsf{T}}}{\sqrt{d_k}}\right) V^{(1)}
\end{align}
Following cross-attention, each branch refines its features through self-attention:
\begin{align}
  SA_v^{(1)} &= \text{Softmax}\left(\frac{CA_{v \to q}^{(1)} CA_{v \to q}^{(1)\mathsf{T}}}{\sqrt{d_k}}\right) CA_{v \to q}^{(1)} \\
  SA_q^{(1)} &= \text{Softmax}\left(\frac{CA_{q \to v}^{(1)} CA_{q \to v}^{(1)\mathsf{T}}}{\sqrt{d_k}}\right) CA_{q \to v}^{(1)}
\end{align}
The outputs of the self-attention stages serve as inputs to the next iteration of the block, thus promoting progressive enhancement of modality alignment. This process continues until the $n$-th layer, after which the refined features of the video and query are output. Therefore, when $1\le i \le n-1$, the iterative expression is given as follows.
\begin{align}
  V^{(i+1)} &= SA_v^{(i)} \\
  Q^{(i+1)} &= SA_q^{(i)}
\end{align}

\section{Details of the VTG head}
\label{VTG}
\subsection{Moment retrieval head}
The design of this head is similar to the foreground head, except it features a last layer with two output channels for the left and right offsets. Given $\tilde{\mathbf{V}}^k \in \mathbb{R}^{L_v\times D}$, this head generates a series of offsets $\{\tilde{m}_i\}_{i=1}^{L_v}$ for each unit. We then define the predicted boundary $\tilde{m}_i$ and the corresponding interval $d_i$ (i.e., $d_i = m_i^s - m_i^e$). For training objectives, we use a combination of smooth L1 loss and generalized IoU loss to optimize the model's performance.
\begin{equation}
    \mathcal{L}_\text{b} = \mathbbm{1}_{f_i=1}  \left[
    \lambda_\text{L1}\mathcal{L}_{\text{SmoothL1}}\left(\tilde{d}_i, {d_i}\right)+
    \lambda_\text{iou} \mathcal{L}_\text{iou}\left( \tilde{m}_i, {m_i} \right) \right].
\end{equation}

Notably, this regression objective is only devised for foreground \unit s~\textit{i.e.,} $f_i=1$. 
\subsection{Video summarization head}
From the frozen video encoder, the output $\tilde{\mathbf{V}}^k\in \mathbb{R}^{L_v\times D}$ passes through a series of three $1\times 3$ convolutional layers, each layer having $D$ filters and equipped with ReLU activation functions. Following these layers, sigmoid activations are used to generate the prediction $\tilde{f}_i$ for each unit. Focal loss serves as the training objective, with $\gamma = 2.0$ and $\alpha = 0.9$.

\begin{equation}
\mathcal{L}_\text{f}=-\lambda_{f} \alpha (1 - \tilde{f}_i)^\gamma \log(\tilde{f}_i)
\label{bce}
\end{equation}
\subsection{Highlight detection head}
Given that saliency is defined as the relevance between visual context and a text query, it is appropriate to assess this relationship through a similarity measure between video and text modalities. Let the video tokens be denoted as $\{\mathbf{v}_i\}_{i=1}^{L_v}\in \mathbb{R}^{L_v\times D}$ and the sentence representation as $\mathbf{S}\in \mathbb{R}^{1\times D}$. We then calculate the predicted saliency score $\tilde{s}_i$ for each video token $\mathbf{v}_i$ in relation to the text query $Q$, using their cosine similarities.
\begin{equation}
\tilde{s}_i = \cos(\mathbf{v}_i, \mathbf{S}) := \frac{\mathbf{v}_i^T\mathbf{S}}{\|\mathbf{v}_i\|_2 \|\mathbf{S}\|_2},
\end{equation}
where $\|\cdot\|_2$ represents the $L2$-norm of a vector.

For each video $\mathbf{V}$, we randomly sample a foreground \unit~$\mathbf{v}_p$ with $f_p=1$ and $s_p>0$ as a positive sample; we treat other clips in the same video $\mathbf{v}_j$ with saliency $s_j$ less than $s_p$ as negative samples, \textit{i.e.,} $\Omega=\{j|s_j<s_p, 1 \leq j \leq {L}_v\}$, and perform \textbf{intra-video} contrastive learning:
\begin{equation}
\small{
\mathcal{L}_\text{s}^\text{intra}=-\log  \frac{\exp \left(\tilde{s}_p/\tau  \right)}{\exp \left(\tilde{s}_p/\tau  \right) + \sum_{j\in \Omega}\exp\left( \tilde{s}_j/ \tau \right)},
\label{intra}
}
\end{equation}
where $\tau$ is a temperature parameter and set as $0.07$. And we further propose query-driven clip-by-clip contrastive learning where clips within the target moment are treated as positive samples and clips outside as negative samples. Specifically, samples are selected based on the salience scores, with positive samples ranked in descending order and negative samples in ascending order. The top \( K \) samples from each are chosen for similarity computation. Given two sets of samples, \( \text{Pos} \) (positive) and \( \text{Neg} \) (negative), each containing \( K \) elements, the similarity is computed using the dot product, resulting in a similarity matrix \( \mathbf{S} \). The similarity matrix \( \mathbf{S} \) is derived from the dot product between vectors \( \mathbf{v}_i^{+} \) from the positive set \( \text{Pos} \) and \( \mathbf{v}_j^{-} \) from the negative set \( \text{Neg} \). Each vector represents a moment in the video, with \( \mathbf{v}_i^{+} \in \text{Pos} \) and \( \mathbf{v}_j^{-} \in \text{Neg} \). The similarity \( S_{ij} \) between any two moments is computed as follows:
\begin{equation}
\label{sim}
 S_{ij} = (\mathbf{v}_i^{+}) \cdot (\mathbf{v}_j^{-})^T,
\end{equation}
The loss function is defined as the negative mean of the trace of \( \mathbf{S} \), formally given by:
\begin{equation}
\label{loss_v}
 \mathcal{L}_\text{v}^\text{intra} = -\frac{1}{N} \sum_{i=1}^{N} \mathbf{S}_{ii}.   
\end{equation}
where $N$ is the clip number of the training set. In datasets other than QVHighlight~\cite{lei2021detecting}, where ground truth salience scores are not provided, the foreground flag \( f \) is used to dichotomize the samples into positive and negative sets. \( K \) samples are then randomly selected from each set for computing the similarity and loss in terms of Eq.~\ref{sim} and \ref{loss_v}.

Besides, we regard sentences from other samples within batches $k\in B$ as negative samples, and develop the \textbf{inter-video} contrastive learning for cross-sample supervision:
\begin{equation}
\mathcal{L}_\text{s}^\text{inter}=-\log\frac{\exp \left( \tilde{s}_p/\tau  \right)}{\sum_{k\in B}  \exp \left(\tilde{s}_p^k / \tau \right)},
\label{inter}
\end{equation}
where $B$ is the training batch size and $\tilde{s}_p^k=\cos (\mathbf{v}_i, \mathbf{S}_k)$.

Our saliency score head training loss is the combination of inter- and intra-video contrastive learning:
\begin{equation}
    \mathcal{L}_\text{s} = \lambda_\text{inter}\mathcal{L}_\text{s}^\text{inter} + \lambda_\text{intra}(\mathcal{L}_\text{s}^\text{intra} + \mathcal{L}_\text{v}^\text{intra}).
\end{equation}

To this end, our grounding objective is the combination of each head loss overall clips in the training set.
\begin{equation}
    \mathcal{L}_{G}=\frac{1}{N}\sum_{i=1}^N\left( \mathcal{L}_\text{f}+\mathcal{L}_\text{b}+\mathcal{L}_\text{s} \right),
\end{equation}
where $N$ is the clip number of the training set.



\end{document}